\documentclass[12pt,onecolumn,draftcls]{IEEEtran}
\usepackage{graphicx}
\usepackage{epsfig}
\usepackage{subfigure}
\usepackage{amssymb}
\usepackage{amsbsy}
\usepackage{amsmath}
\usepackage{steinmetz}

\usepackage{cite}
\usepackage{url}

\usepackage{amsfonts}
\DeclareMathOperator*{\argmin}{\arg\min}  

\usepackage{caption}

\usepackage{color}
\usepackage{float}
\usepackage{times,amsmath,epsfig}
\usepackage{xspace,latexsym,syntonly}
\usepackage{amssymb}
\usepackage{textcomp}

\newtheorem{definition}{Definition}
\newtheorem{theorem}{Theorem}
\newtheorem{remark}{Remark}
\newtheorem{lemma}{Lemma}

\newcommand{\tk}{\boldsymbol{\theta_k}}
\newcommand{\tstar}{\boldsymbol{\theta^\star}}

\begin{document}
\title{On Analog Gradient Descent Learning over Multiple Access Fading Channels}
\author{Tomer Sery, Kobi Cohen
\thanks{Tomer Sery is with the School of Electrical and Computer Engineering, Ben-Gurion University of the Negev, Beer Sheva 8410501 Israel. Email:seryt@post.bgu.ac.il}
\thanks{Kobi Cohen is with the School of Electrical and Computer Engineering, with the Cyber Security Research Center, and with the Data Science Research Center, at Ben-Gurion University of the Negev, Israel. Email: yakovsec@bgu.ac.il}
\thanks{This work has been submitted to the IEEE for possible publication. Copyright may be transferred without notice, after which this version may
no longer be accessible.}
}
\maketitle
\begin{abstract}

We consider a distributed learning problem over multiple access channel (MAC) using a large wireless network. The computation is made by the network edge and is based on received data from a large number of distributed nodes which transmit over a noisy fading MAC. The objective function is a sum of the nodes' local loss functions. This problem has attracted a growing interest in distributed sensing systems, and more recently in federated learning. We develop a novel Gradient-Based Multiple Access (GBMA) algorithm to solve the distributed learning problem over MAC. Specifically, the nodes transmit an analog function of the local gradient using common shaping waveforms and the network edge receives a superposition of the analog transmitted signals used for updating the estimate. GBMA does not require power control or beamforming to cancel the fading effect as in other algorithms, and operates directly with noisy distorted gradients. We analyze the performance of GBMA theoretically, and prove that it can approach the convergence rate of the centralized gradient descent (GD) algorithm in large networks. Specifically, we establish a finite-sample bound of the error for both convex and strongly convex loss functions with Lipschitz gradient. Furthermore, we provide energy scaling laws for approaching the centralized convergence rate as the number of nodes increases. Finally, experimental results support the theoretical findings, and demonstrate strong performance of GBMA using synthetic and real data.
\end{abstract}

\section{Introduction}

We consider a distributed learning problem over a large number of distributed nodes (e.g., sensor nodes, mobile devices, etc.). Specifically, the network consists of $N$ nodes and a network edge (e.g., parameter server in distributed learning systems, fusion center in sensor networks, base station in wireless communications). The objective function is a sum of the nodes' local loss functions. The objective of the network edge is thus to solve the following optimization problem:
\begin{equation}
\label{eq:theta_star_intro}
\boldsymbol{\theta^*} = \argmin_{\boldsymbol{\theta} \in \Theta} \frac{1}{N}\sum_{n=1}^N f_n(\boldsymbol{\theta})
\end{equation}
based on data received from the nodes. The term $\boldsymbol{\theta}\in\Theta\subset\mathbb{R}^d$ is the $d\times 1$ parameter vector which needs to be optimized. The solution $\boldsymbol{\theta^*}$ is known as the empirical risk minimizer. In machine learning tasks, we typically have $f_n(\boldsymbol{\theta})=\ell(\boldsymbol{x}_n, y_n; \boldsymbol{\theta})$, which is the loss of the prediction on input-output data pair sample $(\boldsymbol{x}_n, y_n)$ made with model parameter $\boldsymbol{\theta}$. The goal is to train the algorithm so as to find $\boldsymbol{\theta}$ that transforms the input vector $\boldsymbol{x}$ into the desired output $y$. 

This class of problems have been traditionally solved by centralized GD or Stochastic GD (SGD) algorithms, in which the optimizer has access to each sample directly (e.g., when all data is stored and processed at the cloud). With the increasing demand of data-intensive applications, however, the centralized approach becomes highly inefficient in terms of storage, and latency consumption. Federated learning is a new collaborative machine learning framework suggested to address this issue. In federated learning, the training procedure is distributed among a large number of nodes, each associated with a local loss function. The nodes communicate with the parameter server (PS) that solves (\ref{eq:theta_star_intro}). The problem finds applications in  distributed sensing and control systems as well (see related work in Section \ref{ssec:related}, and numerical examples in Section \ref{sec:sim}). Thus, it is extremely important to develop learning algorithms for these applications which are efficient in terms of communication resources. 

\subsection{Resource-Efficient Communications using MAC}
\label{ssec:intro_from}

The most of existing studies on distributed learning have focused on solving (\ref{eq:theta_star_intro}) using traditional FDM/TDM communication schemes, in which each node sends a function of its observation to the network edge or neighbors using orthogonal channels until convergence (see e.g., \cite{Blatt_convergent_invcremental_const_step, Lopes_Incremental2007, Nedich2007StocInc, ram2009incremental, chen2018lag, konevcny2016federated} and references therein). However, these approaches suffer from highly demanding bandwidth requirements which increase linearly with the number of nodes, and high energy consumption due to the additive noise in each dimension. Furthermore, incremental updates suffer from slow convergence due to cycling messages among nodes. We focus on distributed learning over MAC to overcome these issues. 

Using inference over MAC, each node transmits an analog function of its data over MAC, and the network edge receives a superposition of the analog transmitted signals. The inference decision can be made by the network edge given that the aggregated signals yield a (variation of a) sufficient statistics for the inference task. The number of dimensions used for transmitting the data is independent of $N$, which makes it highly energy and bandwidth efficient.

Analog transmission schemes over MAC have been studied under various inference settings in the sensor network literature (see our previous work on model-dependent inference over MAC \cite{cohen2013performance, cohen2018spectrum}, and references therein, as well as Section \ref{ssec:related}). Although the theoretical performance analysis has been established rigorously under a wide class of problem settings, all these studies assumed that the observation distributions are known to the nodes or to the network edge. Therefore, developing efficient inference algorithms over MAC in the online learning context, where the observation distributions are unknown, becomes extremely important to expand their applicability to real-world problems. 

\subsection{Distributed Learning over MAC}
\label{ssec:intro_distributed}

\emph{Analog transmission schemes over MAC in the online learning context is a new research direction, and very little has been done in this direction so far.} Motivated by the rise of federated learning, this research direction has started to receive a growing attention in the last year. In \cite{amiri2019machine, amiri2019over, amiri2019federated}, the authors developed the compressed analog distributed stochastic gradient descent (CA-DSGD) algorithm, in which each node transmits a sparse parameter gradient vector over MAC. In the case of fading channels \cite{amiri2019federated}, each node uses power control to cancel the channel effect at the receiver, where nodes that experience deep fading do not transmit. In \cite{amiri2019collaborative}, the authors extended the method for transmitting without knowing the channel state at the transmitter. The channel fading is mitigated at the receiver by using multiple antennas, where the fading diminishes as the number of antennas approaches infinity. In \cite{zhu2018low, zhu2019broadband}, the authors considered transmissions over fading MAC, where each entry of the gradient vector is scheduled for transmission depending on
the corresponding channel condition. They developed the federated edge learning (FEEL) algorithm, where each node updates the SGD estimate for multiple steps, and then communicates with the server for model aggregation. Further developments of FEEL used to reduce the energy consumption were developed in \cite{zeng2019energy}. In \cite{yang2018federated}, multiple antennas were used at the receiver, where beamforming was used to maximize the number of devices scheduled for transmission.

\subsection{Main Results}
\label{ssec:main_results}

We focus on developing and analyzing distributed learning over MAC. Below, we summarize our main contributions. \vspace{0.2cm}

\noindent
\textbf{Algorithm Development and Design Parameters:}
We propose a novel Gradient Based Multiple Access (GBMA) algorithm to solve (\ref{eq:theta_star_intro}) over noisy fading MAC. In GBMA, each node transmits an analog function of its local gradient using $d$ common shaping waveforms, one for each element in the gradient vector. The network edge receives a superposition of the analog transmitted signals which represents a noisy (due to the additive noise) distorted (due to the fading channel) version of the global gradient. The network edge updates the estimate and feedbacks the update to the nodes. This procedure continues until convergence (convergence analysis is discussed later). A detailed description of the algorithm is given in Section \ref{sec:GBMA}. 

By using MAC in GBMA, the bandwidth requirement does not depend on $N$, which is a main advantage of inference schemes over MAC. Furthermore, the aggregation of the channel noise is independent of $N$ as well which leads to a significant energy saving. Finally, GBMA uses a GD type learning which does not require complex calculations, or prior knowledge about the sample distributions as required by other model-dependent MAC schemes, such as Likelihood-Based Multiple Access (LBMA) and Type-Based Multiple Access (TBMA) (a discussion of existing methods appears in Section \ref{ssec:related}).\vspace{0.2cm}

The GBMA algorithm is different from the recently suggested learning methods over MAC, detailed in Section \ref{ssec:intro_distributed}, in the following aspects. In terms of communication scheme, the nodes do not use power control or beamforming to cancel the channel gain effect. In GBMA, the estimate is updated based on the noisy distorted gradient directly. The nodes only use phase correction to produce channel gains with nonzero means at the receiver. This scheme captures a more general transmission model, as well as simplifies the system implementation. This type of transmission schemes was proposed and analyzed in past and recent years using model-dependent inference, such as LBMA (where the sum log-likelihood ratio is distorted), and TBMA (where the observation type is distorted) (see our previous work \cite{cohen2013performance, cohen2018spectrum} and references therein, as well as related work in Section \ref{ssec:related}). In this paper, we first develop and analyze this type of transmission scheme in the online learning context, where the global gradient is distorted. In terms of parameter design, the network edge uses a constant stepsize when updating the estimate, which is preferred over diminishing stepsize (as presented in other related studies). We provide specific design principles for the stepsize that guarantee convergence, by taking into account the gradient distortion due to the fading effect. \vspace{0.2cm}

\noindent
\textbf{Performance Analysis:} \emph{Important open questions on gradient-based learning over MAC are whether it can achieve the convergence rate of the centralized GD algorithm, and what are the energy scaling laws for signal transmissions that allow the best possible convergence rate.} In this paper we address these questions. Specifically, we establish a finite-sample bound of the estimation error for both convex and strongly convex loss functions with Lipschitz gradient, and i.i.d. fading channels across nodes and data collections. The error analysis gives a clean expression of how the three terms--the \emph{initial distance}, due to the error in the initial estimate, the \emph{gradient distortion}, caused by the fading channel effect, and the \emph{additive noise} due to the noisy channel--characterize the error bound. Furthermore, we provide specific design principles of the algorithm parameters, and energy scaling laws for approaching the best possible convergence rate obtained by a centralized GD algorithm as $N$ increases. Specifically, the first main theorem considers the case where the loss function is strongly convex. We show that using a constant stepsize in the iterate updates, and setting the transmission energy of each node to $\Omega(N^{\epsilon-2})$, for some $\epsilon>0$, is sufficient to achieve the best possible convergence rate of order $O(c^k)$, where $k$ is the iteration index, for some $0<c<1$, as $N\rightarrow\infty$. The second main theorem relaxes the strongly convex assumption, and considers the case where the loss function is convex. We show that using a constant stepsize in the iterate updates, and setting the transmission energy of each node to $\Omega(N^{\epsilon-2})$, for some $\epsilon>0$, is sufficient to achieve the best possible convergence rate of order $O(1/k)$ in this case, as $N\rightarrow\infty$. These results imply that we can make the total transmission energy consumption in the network be arbitrarily small by increasing the network size, while approaching the best possible centralized convergence rate.

We further evaluate the performance numerically by presenting simulation results of federated learning, and distributed signal processing applications. The simulation results support the theoretical results, and demonstrate strong performance of the GBMA algorithm even when the theoretical conditions are not met.

\subsection{Related Work} 
\label{ssec:related}

Distributed inference problems in wireless networks have attracted much attention in the fields of signal processing in sensor networks and control systems, and more recently in federated learning applications. In past years, the research was focused mainly on model-dependent approaches, where the observation distributions are assumed known, and transmissions over orthogonal channels among nodes. Methods that reduce the number of transmissions by scheduling nodes with better informative observations were developed in \cite{Appadwedula_Decentralized_2007, Patwari_Hierarchical_2003}. More recently, reducing the number of transmissions by ordering transmissions according to the magnitude of the log likelihood ratio was proposed and analyzed in \cite{Blum_Energy_2008, blum2011ordering, zhang2017ordering, sriranga2018energy}. In our previous work, we developed a method that combines both channel state and quality of observations to achieve energy savings \cite{Cohen_Energy_2011}. In \cite{braca2011asymptotically, braca2012single}, asymptotic consistency was shown using only the highest magnitude of the log likelihood ratio. 
However, the bandwidth increases linearly with the number of nodes when using schemes that transmit on orthogonal channels (i.e., dimension per node). 

As explained in Section \ref{ssec:intro_from}, inference schemes over MAC overcome this issue. Well known transmission schemes for inference over MAC are Likelihood Based Multiple Access (LBMA) (see \cite{Liu_Type_2007, Marano_Likelihood_2007} and our previous work \cite{cohen2013performance}), and Type-Based Multiple Access (TBMA) (see \cite{mergen2006type, Mergen_Asymptotic_2007, Liu_Type_2007}). In LBMA, each node computes the log-likelihood ratio locally, and then amplifies the transmitted waveform by this value. In TBMA, the observations are quantized before communication to $K$ possible levels. Nodes that observe level $k$ transmit a corresponding waveform $k$ from a set of $K$ orthonormal waveforms. The network edge receives a superposition of the waveforms over MAC which allows to make inference decisions. In our very recent work we developed an energy and spectrum efficient improved method \cite{cohen2018spectrum}. Other related works have investigated inference over MAC for using multiple antennas at the network edge \cite{nevat2014distributed}, detection with a non-linear sensing behavior \cite{zhang2016event}, using non-coherent transmissions  \cite{anandkumar2007type, li2011decision}, and detecting a stationary random process distributed in space and time with a circularly-symmetric complex Gaussian distribution \cite{maya2015optimal, maya2015exploiting}. However, all these studies assume that the observation distributions are known to the nodes or to the network edge, which are assumed unknown in this paper. A detailed discussion of analog transmission schemes over MAC in the online learning context, where the observation distributions are unknown was given in Section \ref{ssec:intro_distributed}.

Popular traditional methods for distributed inference in the online learning context use incremental updates among nodes \cite{Blatt_convergent_invcremental_const_step, Lopes_Incremental2007, Nedich2007StocInc, wang2019adaptive, skatchkovsky2019optimizing, Mahmud_incremental2018}. In recent years, other stochastic gradient-descent (SGD) based methods were developed for federated learning \cite{wang2019adaptive, skatchkovsky2019optimizing}. While these methods do not require prior knowledge of the observation distributions, they use orthogonal channels among node transmissions, which results in high bandwidth and energy consumption.  Using MAC for online learning as considered in this paper overcomes these issues.\vspace{0.2cm}

\emph{Notations:} Throughout the paper, all vectors are considered to be column vectors. We denote vectors by boldface lowercase letters, and matrices by boldface uppercase letters. 

\section{System Model and Problem Statement}
\label{sec:system}

We consider a wireless network consisting of $N$ nodes indexed by the set $\mathcal{N} = \{1,2,...,N\}$ and a network edge. As detailed and motivated in the Introduction, each node is associated with a local loss function $f_n$, and the objective function is a sum of the nodes' local loss functions:
\begin{equation}
\label{eq: F definition}
F(\boldsymbol{\theta})\triangleq\frac{1}{N} \sum_{n=1}^N f_n(\boldsymbol{\boldsymbol{\theta}}).
\end{equation} 
The objective of the network edge is to solve the following optimization problem:
\begin{equation}
\label{eq:theta_star}
\boldsymbol{\theta^*} = \argmin_{\boldsymbol{\theta} \in \Theta} \;F(\boldsymbol{\theta})
\end{equation}
based on data received from the nodes. We assume that $f_n$ is convex, and has Lipschitz gradient with Lipschitz constant $L_n$ (see Section \ref{sec:performance} for more details). We denote the maximal Lipschitz constant among all nodes by $\overline{L}\triangleq \max_{n}L_n$. The term $\boldsymbol{\theta}\in\Theta\subset\mathbb{R}^d$ is the $d\times 1$ parameter vector which needs to be optimized. It is assumed that the parameter lies in the interior of a compact convex parameter set $\Theta$, with diameter $\delta$. The solution $\boldsymbol{\theta^*}$ is known as the empirical risk minimizer. 
Each node $n$ is aware only of its local loss function $f_n$, and we denote the gradient of $f_n$ with respect to the unknown parameter at some parameter value $\boldsymbol{\theta'}$ by 
\begin{equation}
    \label{eq: g_n definition}
    \boldsymbol{g_n}(\boldsymbol{\theta'}) = \boldsymbol{\nabla} f_n(\boldsymbol{\theta'}).
\end{equation}

\section{Gradient-Based Learning over MAC}
\label{sec:GBMA}

\begin{figure} [htbp]
\begin{center}
    \includegraphics[height=9cm]{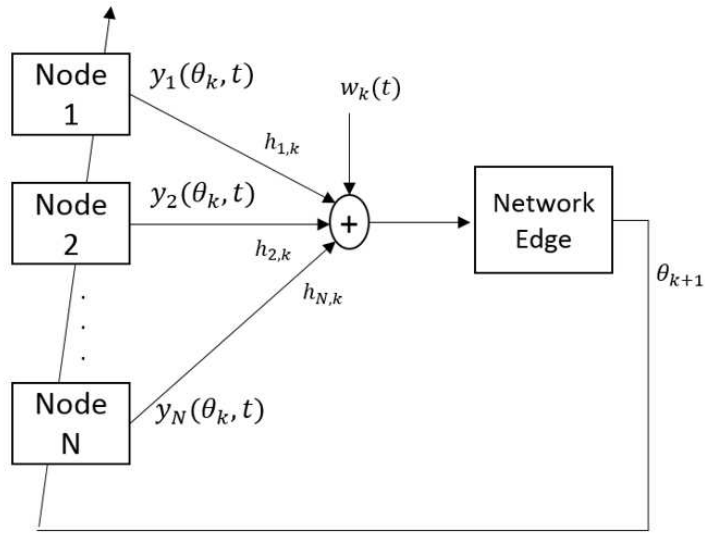}
   \caption{An illustration of the transmission scheme considered in this paper. $y_n(\boldsymbol{\theta_k},t)\triangleq\sqrt{E_N}e^{-j\phi_{n, k}}\boldsymbol{g_n}(\boldsymbol{\theta_k})^T \boldsymbol{s(t)}$ represents the signal transmitted by node $n$ at iteration $k$ (see (\ref{eq:transmitted_signal})). The network edge receives a noisy superposition of the transmitted signals, each multiplied by a random channel gain $h_{n,k}$. After matched-filtering, it generates a new estimate $\boldsymbol{\theta_{k+1}}$ (see (\ref{eq:GD})) and transmits it back to the nodes, which update their local gradients $\boldsymbol{g_n}(\boldsymbol{\theta_{k+1}})$ for the next iteration $t_{k+1}$. This procedure continues until convergence.
   }    \label{fig:scheme} 
\end{center}
  \end{figure}

We now present the GBMA algorithm. Under GBMA, all nodes transmit a function of the local gradient to the network edge simultaneously using common analog waveforms. The network edge updates the estimate based on the received data and feedbacks the updated estimate to the nodes, and so on until convergence. We next discuss the transmission scheme in details. An illustration is given in figure \ref{fig:scheme}. 

Let $\boldsymbol{s(t)}=\left(s_1(t), ..., s_d(t)\right), \;0<t<T$, be a vector of $d$ orthogonal baseband equivalent normalized waveforms, satisfying $\int_{0}^{T}{s_m^2(t)dt}=1$, $\int_{0}^{T}{s_m(t)s_r(t)dt}=0$, for $m\neq r$. The time is slotted, and indexed by $t_1, t_2, ...$. Each node $n$ experiences at time $t_k$ a block fading channel $\tilde{h}_{n,k}$ with gain $h_{n,k}\triangleq|\tilde{h}_{n,k}|\in \mathbb{R}_+$ and phase $\phi_{n,k}\triangleq\phase{\tilde{h}_{n,k}}\in \left\{x\in\mathbb{R}|-\pi\leq x\leq\pi\right\}$. The channel fading is assumed i.i.d. across nodes, and time slots, with mean $\mu_h$ and variance $\sigma_h^2$. The local channel state is assumed known for each node, which is typically estimated by receiving a beacon signal before transmission \cite{Mergen_Asymptotic_2007, cohen2013performance, wimalajeewa2015wireless}. Each time slot $t_k$ is divided into two phases, and is associated with a single iterate of the GBMA algorithm. Let $\boldsymbol{\theta_k}$ be the updated estimate of the parameter at iteration $k$. In the first phase, all nodes transmit a linear combination of $d$ amplified orthogonal analog signals\footnote{In typical OFDM systems, we can transmit $d\approx 10^5$ orthogonal waveforms with $T$ smaller than the channel coherence time ($\approx2-5$ms). For larger problem dimensions (say $\approx r\cdot 10^5$), we can transmit the data over $r$ consecutive time slots, and the analysis applies with slight modifications.}:
\begin{equation}
\label{eq:transmitted_signal}
\sqrt{E_N}e^{-j\phi_{n, k}}  \boldsymbol{g_n}(\boldsymbol{\theta_k})^T \boldsymbol{s(t)}, \end{equation}
where $E_N$ is set to satisfy the energy requirement, and $e^{-j\phi_{n, k}}$ is due to phase correction at the receiver as suggested in past studies (e.g., \cite{Mergen_Asymptotic_2007, cohen2013performance, wimalajeewa2015wireless}). It should be noted that phase correction is only needed to produce channel gains with nonzero means at the network edge. In the case where the channel gains have nonzero means, phase correction is not required. In the case where the channel gains have zero mean, correcting the phase with an error less than $\pi/4$ is sufficient to yield channel gains with nonzero means at the receiver. Therefore, only partial information about the channel phase is required. The network edge receives a superposition of all transmitted signals:
\begin{equation}
\begin{array}{c}
 \displaystyle r_k(t) = \sum_{n=1}^{N}\sqrt{E_N}  h_{n,k} \boldsymbol{g_n}(\boldsymbol{\theta_k})^T \boldsymbol{s(t)}+ w_k(t),\vspace{0.2cm}\\\hspace{5cm}
 \displaystyle
 t_k\leq t <t_k+T,
\end{array}
\end{equation}
where $w_k(t)$ is a zero-mean additive Gaussian noise process at time $t_k$. After matched-filtering $r_k(t)$ by the $d$ corresponding waveforms at the network edge, we get the following $d\times 1$ projected signal vector:
\begin{equation}
\boldsymbol{\tilde{v}_k} = \sum_{n=1}^{N} \sqrt{E_N} h_{n,k} \boldsymbol{g_n}(\boldsymbol{\theta_k}) + \boldsymbol{\tilde{w}_k},
\end{equation}
where element $m$ of $\boldsymbol{\tilde{v}_k}$ corresponds to the projection of $r_k(t)$ on $s_m(t)$. The term $\boldsymbol{\tilde{w}_k}$ is a zero-mean additive Gaussian noise vector, distributed as $\boldsymbol{\tilde{w}}_k\sim N(0,\sigma_w^2 \boldsymbol{I_d})$, where $\boldsymbol{I_d}$ is the $d\times d$ identity matrix.
Let
\begin{equation} 
\label{eq:v_k definition}
\boldsymbol{v_k} \triangleq \frac{\boldsymbol{\tilde{v}_k}}{N \sqrt{E_N}} = \frac{1}{N} \sum_{n=1}^{N}  h_{n,k} \boldsymbol{g_n}(\boldsymbol{\theta_k}) + \boldsymbol{w_k},
\end{equation}
where $\boldsymbol{w_k}\triangleq\frac{\boldsymbol{\tilde{w}_k}}{N\sqrt{E_N}}\sim N(0,\frac{\sigma_w^2}{N^2E_N} \boldsymbol{I_d})$. 
Then, the network edge updates the estimate $\boldsymbol{\theta_{k+1}}$ by a GD-type iterate:
\begin{equation} 
\label{eq:GD}
\boldsymbol{\theta_{k+1}} = \boldsymbol{\theta_k} - \beta \boldsymbol{v_k}, 
\end{equation}
for $k=0, 1, ...$, where $\boldsymbol{\theta}_0$ is the initial estimate of the parameter. The term $\boldsymbol{v_k}$ represents a noisy distorted version of the global gradient of $F(\boldsymbol{\theta_k})$, and $\beta$ is a constant stepsize that will be designed later.   

In the second phase of time slot $t_k$, the network edge broadcasts the estimate $\boldsymbol{\theta_{k+1}}$ back to the nodes, which update their local gradients $\boldsymbol{g_n}(\boldsymbol{\theta_{k+1}})$ for the next iteration $t_{k+1}$. This procedure continues until convergence.\vspace{0.2cm}

\begin{remark}
Note that in the case of noiseless channel, and equal channel gains, $\boldsymbol{v_k}$ represents the true global gradient. Thus, the GBMA algorithm achieves the convergence rate of the centralized GD algorithm. \emph{We are thus interested in analyzing the performance of the distributed GBMA algorithm over MAC in the noisy fading channel setting considered in this paper, and design energy scaling laws to achieve the best possible convergence rate.}
\end{remark}

\subsection{Implementation Discussion of the GBMA Algorithm}
The implementation of GBMA has important advantages for inference tasks using wireless networks. It is highly bandwidth efficient since the bandwidth requirement is independent of the number of nodes, in contrast to TDM/FDM schemes in which the bandwidth requirement increases linearly with the number of nodes. Furthermore, the aggregated channel noise is independent of the number of nodes which leads to a significant energy saving as compared to TDM/FDM schemes. Second, GBMA does not require prior knowledge of the observation distributions, as required by well known inference methods, such as LBMA, and TBMA (see section \ref{ssec:related} for more details).

\section{Preliminaries}

In Section \ref{sec:performance} we will analyze the convergence rate of the GBMA algorithm. In this section we provide a background on definitions and lemmas used in the optimization and learning literature that will be used throughout the analysis (for more details on the background provided in this section the reader is referred to \cite{Nesterov1998CovexProg}). We start by defining the commonly used linear and sublinear convergence rates with the number of iterations used in GD-based learning algorithms, which intuitively speaking are motivated by linear and sublinear curves, respectively, on a semi-log plot.\vspace{0.2cm}

\noindent
\begin{definition}
If an algorithm converges with rate $O(c^k)$, for $0<c<1$, and $k$ is the number of iterations, this rate is referred to as \emph{linear convergence}. If the algorithm converges with rate $O(1/k)$, this rate is referred to as \emph{sublinear convergence}. \vspace{0.2cm}
\end{definition}

Next, we define functions with $L$-Lipschitz continuous gradient.\vspace{0.2cm}

\noindent
\begin{definition}
A function $f(\boldsymbol{x})$ with domain $X$ has a Lipschitz continuous gradient if it is continuously differentiable for any $\boldsymbol{x}\in X$, and the inequality
\begin{equation}
\displaystyle ||\boldsymbol{\nabla} f(\boldsymbol{x}) - \boldsymbol{\nabla} f(\boldsymbol{y})|| \leq L||\boldsymbol{x}-\boldsymbol{y}||
\end{equation}
holds for all $\boldsymbol{x},\boldsymbol{y}\in X$. The constant $L$ is called the Lipschitz constant.\vspace{0.2cm}
\end{definition}

Finally, we define the strong convexity property. \vspace{0.2cm}

\noindent
\begin{definition}
A function $f(\boldsymbol{x})$ with domain $X$ is $\mu$-strongly convex if it is continuously differentiable for any $\boldsymbol{x} \in X$ and the inequality
\begin{equation}
\displaystyle \left<\boldsymbol{\nabla} f(\boldsymbol{x}) - \boldsymbol{\nabla} f(\boldsymbol{y}), \boldsymbol{x}-\boldsymbol{y} \right> \geq \mu||\boldsymbol{x}-\boldsymbol{y}||^2
\end{equation}
holds for all $\boldsymbol{x},\boldsymbol{y}\in X$. The constant $\mu$ is called the strong convexity constant.\vspace{0.2cm}
\end{definition}

Below, we present useful lemmas of the linearity of strong convexity and Lipschitz continuous properties that will be used in the analysis.\vspace{0.2cm}

\noindent
\begin{lemma}
Consider two Lipschitz continuous functions, $f(\boldsymbol{x}), g(\boldsymbol{x})$ with Lipschitz constants $L_f$ and $L_{g}$, 
respectively. Then, the function $h(\boldsymbol{x})= \alpha f(\boldsymbol{x}) +\beta g(\boldsymbol{x})$ is Lipschitz continuous with Lipschitz constant $L_h =\alpha L_{f}+ \beta L_{g}$.\vspace{0.2cm}
\end{lemma}

\noindent
\begin{lemma}
Consider two strongly convex functions, $f(\boldsymbol{x}), g(\boldsymbol{x})$ with constants $\mu_f$ and $\mu_{g}$, respectively. Then, the function $h(\boldsymbol{x})= \alpha f(\boldsymbol{x}) +\beta g(\boldsymbol{x})$ is strongly convex with constant  $\mu_h =\alpha \mu_{f}+\beta \mu_{g}$.\vspace{0.2cm}
\end{lemma}

\begin{remark}
In this paper we are interested in analyzing an objective function $F(\boldsymbol{\theta})$ which is the average over local functions 
$f_n(\boldsymbol{\theta})$ (see \eqref{eq: F definition}). Therefore, if we assume $\mu_n$-strong convexity of $f_n(\boldsymbol{\theta})$, this implies $\left(\frac{1}{N}\sum_{n=1}^N \mu_{n}\right)$-strong convexity of $F(\boldsymbol{\theta})$. Similarly, if we assume $L_n$-Lipschitz gradient of $f_n(\boldsymbol{\theta})$, this implies   $\left(\frac{1}{N}\sum_{n=1}^N L_{n}\right)$-Lipschitz gradient of $F(\boldsymbol{\theta})$.\vspace{0.2cm}
\end{remark}

\noindent
\begin{lemma}
Let $f(\boldsymbol{x})$ denote a $\mu$-strongly convex function with $L$- Lipschitz gradient. Then, the following inequality holds:
\begin{multline}
\label{eq: 2.1.15 inequality}
\left < \boldsymbol{\nabla} f(\boldsymbol{x}) - \boldsymbol{\nabla} f(\boldsymbol{y}), \boldsymbol{x}-\boldsymbol{y} \right > \geq \frac{\mu L}{\mu+L}||\boldsymbol{x}-\boldsymbol{y}||^2 \\ +\frac{1}{\mu+L}||\boldsymbol{\nabla} f(\boldsymbol{x}) - \boldsymbol{\nabla} f(\boldsymbol{y})||^2.\vspace{0.2cm}
\end{multline}
\end{lemma}

\begin{lemma}
\label{lemma:lip_ineuality}
Let $f(\boldsymbol{x})$ denote a convex function with $L$-Lipschitz gradient. Then, the following inequality holds:
\begin{equation}
\label{eq:2.1.7}
\begin{array}{l}
\displaystyle\frac{1}{2L}||\boldsymbol{\nabla} f(\boldsymbol{x}) -\boldsymbol{\nabla} f(\boldsymbol{y})||^2
\leq f(\boldsymbol{y}) - f(\boldsymbol{x}) - \left<\boldsymbol{\nabla} f(\boldsymbol{x}), \boldsymbol{y}-\boldsymbol{x} \right>
\vspace{0.2cm}\\\hspace{6cm}
\displaystyle\leq \frac{L}{2}||\boldsymbol{x}-\boldsymbol{y}||^2.
\end{array}
\end{equation}
\end{lemma}
The proofs for the lemmas in this section can be found in \cite{Nesterov1998CovexProg}.

\section{Performance Analysis}
\label{sec:performance}

In this section, we analyse the performance of the GBMA algorithm. The index $n$ is used for the node index, and $k$ is used for the iterate update at time slot $t_k$. The error (or the excess risk) of GD type algorithms is commonly defined as the loss in the objective value at iteration $k$ with respect to the optimal value:
\begin{equation}
\label{eq:excess_risk}
\displaystyle\mathbb{E}[F(\boldsymbol{\theta_k})]-F(\boldsymbol{\theta^*}),
\end{equation}
where the expectation is over the estimator $\boldsymbol{\theta_k}$.

We are interested in characterizing the rate at which the error decreases with the number of iterations $k$. Furthermore, since the error depends on the number of nodes and the transmission energy as well in our distributed MAC setting (as detailed in the analysis), we are interested in establishing energy scaling laws for signal transmissions used to achieve the best possible convergence rate order. 

\subsection{Analyzing the Case of Strongly Convex Objective Function with Lipschitz Gradient}
\label{ssec:strongly_lip}

In this section we analyze the performance of the GBMA algorithm under the assumption that $F(\boldsymbol{\theta})$ is strongly convex and has Lipschitz gradient. The centralized GD algorithm is known to achieve linear convergence rate under strongly convex with Lipschitz gradient functions. In the main theorem below we establish a finite-sample bound of the error for finite $k$, and $N$. In Section \ref{ssec:discussion}, we will discuss the energy scaling laws for signal transmissions that guarantee linear convergence rate as $N$ increases. \vspace{0.2cm}

\begin{theorem}
\label{th:error_bound_fading_channel}
Consider the system model specified in Section \ref{sec:system}. Let $\boldsymbol{\theta^*}$ denote the solution of the optimization problem in (\ref{eq:theta_star}). Let $\mu, L$ be the strong convexity, and Lipschitz gradient constants of $F(\boldsymbol{\theta})$, respectively. Let $r_0^2 \triangleq ||\boldsymbol{\theta}_0 -\boldsymbol{\theta^*}||^2$
be the squared distance between the initial estimate $\boldsymbol{\theta}_0$ and $\boldsymbol{\theta^*}$. Let the constant stepsize in \eqref{eq:GD} satisfy:
\begin{equation}
    \label{eq:step_condition_strong_convex_gradient_unbounded}
    \displaystyle 0 < \beta < \min \left\{ \frac{2}{\mu_h(\mu+L)},\frac{2 \mu_h \mu L N}{\sigma_h^2 \overline{L}^2(1+2\delta)  (\mu+L)} \right\}.
\end{equation}
Then, the error under GBMA is bounded by:
\begin{equation}
\label{eq:th_error_f_equal_strong_convex}
\begin{array}{l}
\displaystyle 
\mathbb{E}[F(\boldsymbol{\theta_k})]-F(\boldsymbol{\theta^*}) \vspace{0.2cm}\\\hspace{0.3cm}
\displaystyle \leq  c^k r_0^2\frac{L}{2} +\frac{L\beta^2}{2(1-c)}\left(\frac{\sigma_h^2 \delta\overline{L}^2(2+\delta)}{N} +\frac{d\sigma_w^2} {E_N N^2}\right),
\end{array}
\end{equation}
where $\displaystyle 0<c\triangleq 1-\frac{2\beta \mu_h\mu L}{\mu +L} + \frac{\beta^2\sigma_h^2 \overline{L}^2(1+2\delta)}{N}<1$.\vspace{0.2cm} \end{theorem}

The proof is given in Appendix \ref{app:proof1}. A detailed discussion of the results is given in Section \ref{ssec:discussion}. \vspace{0.2cm}

\subsection{Analyzing the Case of Convex Objective Function with Lipschitz Gradient}
\label{ssec:convex_lip}

In this section we relax the assumption of strongly convex objective function, and assume that $F(\boldsymbol{\theta})$ is convex. We still assume that $F(\boldsymbol{\theta})$ has Lipschitz gradient. The centralized GD algorithm is known to achieve a sublinear convergence rate order of $O(1/k)$ under convex with Lipschitz gradient functions. In the main theorem below, we establish a finite-sample bound of the error, for finite $k$, and $N$. In Section \ref{ssec:discussion}, we will discuss the energy scaling laws for signal transmissions that guarantee the sublinear convergence rate of $O(1/k)$ as $N$ increases. \vspace{0.2cm}

\begin{theorem}
\label{th:error_bound_fading_channel_convex}
Consider the system model specified in Section \ref{sec:system}. Let $\boldsymbol{\theta^*}$ denote the solution of the optimization problem in (\ref{eq:theta_star}). Assume that $F(\boldsymbol{\theta})$ is convex with Lipschitz gradient, and let $L$ be its Lipschitz gradient constant. Assume that 
\begin{equation}
\label{eq:condition_decreasing}
\displaystyle \mathbb{E}\left[ || \boldsymbol{\nabla} F(\boldsymbol{\theta}_i)||^2\right] >  \frac{d\sigma_w^2}{E_N N^2}, \;\;\mbox{for all} \;\;i=1, ..., k.
\end{equation}
Let $r_0^2 \triangleq ||\boldsymbol{\theta}_0 -\boldsymbol{\theta^*}||^2$
be the squared distance between the initial estimate $\boldsymbol{\theta}_0$ and $\boldsymbol{\theta^*}$. Then, the following statements hold: \vspace{0.2cm}\\
a) (The case of equal channel gains:) Assume that $h_{n,k}=1$ for all $n, k$. Let the constant stepsize in \eqref{eq:GD} satisfy:
\begin{equation} 
\label{eq: beta condition_equal_convex}
\displaystyle 0 < \beta < \frac{1}{L}.
\end{equation}
Then, the error under GBMA is bounded by:
\begin{equation}
\label{eq:th_error_f_equal_convex}
\displaystyle 
\mathbb{E}[F(\boldsymbol{\theta_k})]-F(\boldsymbol{\theta^*}) \leq \frac{r_0^2}{2\beta k}+\frac{\beta d\sigma_w^2} {E_N N^2}.\vspace{0.2cm}
\end{equation}
b) (The case of fading channels:) Assume that the expectation of $||\boldsymbol{\nabla} f_n(\boldsymbol{\theta})||^2$ satisfies: $\mathbb{E}\left[||\boldsymbol{\nabla} f_n(\boldsymbol{\theta})||^2\right]\leq B(N)$ for all $\boldsymbol{\theta}$ and $n$, where $B(N)$ is a function of $N$. Let the constant stepsize in \eqref{eq:GD} satisfy:
\begin{equation} 
\label{eq: beta condition_fading_convex}
\displaystyle 0 < \beta < \frac{1}{L\mu_h}.
\end{equation}
Then, the error under GBMA is bounded by:
\begin{equation}
\label{eq:th_error_f_fading_convex}
\begin{array}{l}
\displaystyle 
\mathbb{E}[F(\boldsymbol{\theta_k})]-F(\boldsymbol{\theta^*}) \leq \frac{r_0^2}{2\beta\mu_h k}+\frac{\beta}{\mu_h}\left(\frac{B(N)\sigma_h^2}{N}+\frac{d\sigma_w^2} {E_N N^2}\right).\vspace{0.2cm}
\end{array}
\end{equation}
\end{theorem}

The proof is given in Appendix \ref{app:proof2}. A detailed discussion of the results is provided in the next section.

\subsection{Discussion on the Main Theorems \ref{th:error_bound_fading_channel}, and \ref{th:error_bound_fading_channel_convex}}
\label{ssec:discussion}

We now provide important insights about the convergence rate of the GBMA algorithm implied by Theorems \ref{th:error_bound_fading_channel}, and \ref{th:error_bound_fading_channel_convex}. \vspace{0.2cm}

\subsubsection{\textbf{Characterization of the error bound}}
Theorems \ref{th:error_bound_fading_channel}, and \ref{th:error_bound_fading_channel_convex} give a clean expression of how the three terms--the \emph{initial distance}, due to the error in the initial estimate, the \emph{gradient distortion}, caused by amplifying each local gradient by a different random channel gain, and the \emph{additive noise} due to the noisy channel--characterize the error bound. \vspace{0.2cm}

In a centralized GD algorithm, the convergence rate is affected by the initial distance only. Specifically, the centralized GD algorithm is known to achieve linear convergence rate:
\begin{equation}
\displaystyle\mathbb{E}[F(\boldsymbol{\theta_k})]-F(\boldsymbol{\theta^*})\leq \left(1-\frac{2\beta\mu L}{\mu +L}\right)^k r_0^2\frac{L}{2}\;,   
\end{equation}
under strongly convex with Lipschitz gradient functions when using constant stepsize $0<\beta<\frac{2}{\mu+L}$. 
Also, it is known to achieve sublinear convergence rate: 
\begin{equation}
\displaystyle\mathbb{E}[F(\boldsymbol{\theta_k})]-F(\boldsymbol{\theta^*})\leq\frac{r_0^2}{2\beta k}\;,
\end{equation}
under convex with Lipschitz gradient functions when using constant stepsize $0<\beta<\frac{1}{L}$.

The terms $c^k r_0^2\frac{L}{2}$ in Theorem \ref{th:error_bound_fading_channel}, and $\frac{r_0^2}{2\beta\mu_h k}$ in Theorem \ref{th:error_bound_fading_channel_convex} clearly explain the connection to the convergence rate with the number of iterations that can be achieved by the centralized GD algorithm. In the case of noiseless channel and equal channel gains, the GBMA algorithm uses the same data as in the centralized GD algorithm, and thus achieves the same performance. The error bounds in Theorems \ref{th:error_bound_fading_channel}, \ref{th:error_bound_fading_channel_convex}, coincide with this observation by setting $\mu_h=1, \sigma_h^2=0, \sigma_w^2=0$. Note that the stepsizes $0 < \beta < \frac{2}{\mu_h (\mu+L)}$ in Theorem \ref{th:error_bound_fading_channel}, and $0 < \beta < \frac{1}{L\mu_h}$ in Theorem \ref{th:error_bound_fading_channel_convex} are intuitively satisfying. In the case of no distortion, i.e., $\sigma_h^2=0$, the global gradient is amplified by $\mu_h$. Thus, normalizing the stepsize by $\mu_h$ is needed. 

Next, we discuss the effect of the gradient distortion on the error bound. Let 
\begin{equation}
\displaystyle D\triangleq\frac{\sigma_h^2}{\mu_h}
\end{equation}
denote the channel index of dispersion, which measures the distortion of the global gradient. The stepsize $\beta$ in Theorem \ref{th:error_bound_fading_channel} is given by:
\begin{equation}
\displaystyle 0< \beta < \min \left\{ \frac{2}{\mu_h(\mu+L)},\frac{2\mu L N}{\overline{L}^2(1+2\delta)  (\mu+L)}\cdot\frac{1}{D} \right\}.
\end{equation}
Thus, the stepsize $\beta$ decreases with $D$. Also, note that the additional term $\frac{\beta^2\sigma_h^2 \overline{L}^2(1+2\delta)}{N}$ in the coefficient rate $c$ in Theorem \ref{th:error_bound_fading_channel} decelerates the linear convergence rate under the strongly convex case. Nevertheless, as $N\rightarrow\infty$, the distortion effect diminishes, and GBMA approaches the linear convergence rate of the centralized GD algorithm. 
Under the convex case, the distortion effect can be viewed by the term $\frac{\beta}{\mu_h}\cdot\frac{B(N)\sigma_h^2}{N}$ in Theorem \ref{th:error_bound_fading_channel_convex}, which  can be rewritten as:
\begin{equation}
\displaystyle\beta\cdot\frac{B(N)}{N}\cdot D,
\end{equation}
and increases with $D$, as expected. 
Next, assume that $B(N)=O(N^{1-\epsilon})$, for some $\epsilon>0$. Then, we have:
\begin{center}
$\displaystyle\beta\cdot\frac{B(N)}{N}\cdot D\rightarrow 0\;\;, \;\;\mbox{as}\;\;  N\rightarrow\infty$.
\end{center}
Thus, in contrast to classic SGD studies that assumed bounded expected gradients to converge (see e.g., \cite{shalev2011pegasos, hazan2014beyond, nemirovski2009robust}), this result implies that we can allow the expected squared gradients be arbitrarily large. The error bound of GBMA approaches the error bound of the centralized GD algorithm by increasing $N$ and consequently diminishing the distortion term.\vspace{0.2cm}

The noise terms, 
$\frac{L\beta^2}{2(1-c)}\cdot\frac{d\sigma_w^2} {E_N N^2}$ in Theorem \ref{th:error_bound_fading_channel}, and $\frac{\beta}{\mu_h}\cdot\frac{d\sigma_w^2}{E_N N^2}$ in Theorem \ref{th:error_bound_fading_channel_convex}, are related to the additive noise, and are affected by the transmitted energy. It is intuitive that the effect of the additive noise increases with $\sigma_w$ and the dimension $d$, and decreases with $N$ and $E_N$. We next establish the energy scaling laws for controlling the noise term.\vspace{0.2cm}

\subsubsection{\textbf{Energy scaling laws for approaching the centralized convergence rate}} 
The transmission energy consumed by the nodes controls the noise terms $\frac{L\beta^2}{2(1-c)}\cdot\frac{d\sigma_w^2} {E_N N^2}$ in Theorem \ref{th:error_bound_fading_channel}, and $\frac{\beta}{\mu_h}\cdot\frac{d\sigma_w^2}{E_N N^2}$ in Theorem \ref{th:error_bound_fading_channel_convex}. The theorems imply that by setting $E_N=\Omega\left(N^{\epsilon -2}\right)$, for some $\epsilon>0$, we get 
\begin{center}
$\displaystyle\frac{L\beta^2}{2(1-c)}\cdot\frac{d\sigma_w^2} {E_N N^2}\rightarrow 0\;\;, \;\;\mbox{as}\;\;  N\rightarrow\infty$,
\end{center}
in Theorem \ref{th:error_bound_fading_channel}, and 
\begin{center}
$\displaystyle\frac{\beta}{\mu_h}\cdot\frac{d\sigma_w^2} {E_N N^2}\rightarrow 0\;\;, \;\;\mbox{as}\;\;  N\rightarrow\infty$,
\end{center}
in Theorem \ref{th:error_bound_fading_channel_convex}. 
This implies that GBMA achieves the centralized convergence rate as $N\rightarrow\infty$ (when  $B(N)=O(N^{1-\epsilon})$ in Theorem \ref{th:error_bound_fading_channel_convex}). Furthermore, this result implies that the estimator's performance can be improved by increasing the number of nodes which participate in the inference task while making the total transmission energy in the network arbitrarily close to zero, by setting $N^{\epsilon-2}\lesssim E_N\lesssim N^{-\epsilon-1}$. These results provide important energy scaling laws for distributed learning in edge computing systems under resource constraints.

It is worth noting that condition (\ref{eq:condition_decreasing}) in Theorem \ref{th:error_bound_fading_channel_convex} is required for technical reasons when proving the theorem (see Appendix \ref{app:proof2}). Note that for any finite $k$ (thus, the expected squared gradient is strictly greater than zero), and by setting $E_N=\Omega\left(N^{\epsilon -2}\right)$, we can choose a sufficiently large $N$ so that condition (\ref{eq:condition_decreasing}) is satisfied.\vspace{0.2cm}

\subsubsection{\textbf{Comparison with SGD based algorithms}} We point out that SGD based algorithms use noisy gradients as well in the algorithm iterates by computing the gradient based on a small number of samples at each iteration. However, SGD achieves $O(1/k)$ convergence rate for strongly convex functions with Lipschitz gradient, and $O(1/\sqrt{k})$ convergence rate for convex functions with Lipschitz gradient. In addition, it requires to use a decreasing stepsize to converge (see a detailed discussion on existing SGD algorithms in our previous work \cite{cohen2017projected} and references therein). By contrast, in this paper, by exploiting the inherent structure of the noisy distorted gradient over MAC used in the GBMA iterates, we overcome the performance reduction occurs by using noisy gradients as in SGD algorithms. Specifically, the GBMA algorithm with constant stepsize approaches the $O(c^k)$, and $O(1/k)$ convergence rate order under strongly convex, and convex functions, respectively, with Lipschitz gradient, as $N$ increases.

\section{Experiments}
\label{sec:sim}

In this section we provide numerical examples to illustrate the performance of GBMA in two different settings. In the first setting, we simulated a federated learning task used to predict a release year of a song from its audio features. The training was distributed among a large number of devices that collaborate to train the predictor. We used real-data, the popular Million Song Dataset, to demonstrate the performance of the algorithm. In the second setting, we focused on a distributed learning task for estimation in sensing systems. Specifically, we demonstrated the performance of the algorithm in a source localization problem using wireless sensor networks. 

\subsection{Federated Learning over the Million Song Dataset}
\label{ssec:real}

we start by examining the performance of the algorithm for prediction of a release year of a song from audio features using a federated learning setting. We used the dataset available by UCI Machine Learning Repository \cite{Lichman:2013}, extracted from the Million Song Dataset collaborative project between The Echo Nest and LabROSA \cite{Bertin-Mahieux2011}. The Million Song Dataset contains songs which are mostly western, commercial tracks ranging from 1922 to 2011. Each song is associated with a released year (i.e., $y$ in our model that we aim to predict), and $90$ audio attributes (i.e., $\boldsymbol{x}$ in our model).

In such dedicated apps, the federated learning approach reduces the storage, and latency consumption at data centers, since each smart device can process a small amount of songs, extract the required features, compute the local gradient, and then collaborate with a large amount of devices to train the predictor. To model this setting, we simulated a network, in which each $(\boldsymbol{x}_n, y_n)$ data pair sample from the Million Song Dataset is stored at a different mobile device. We used a regularized linear least squares loss for each device (say $n$):
\begin{equation}
f_n(\boldsymbol{\theta}) = \frac{1}{2}(\boldsymbol{x_n}^T\boldsymbol{\theta}-y_n)^2 + \frac{\lambda}{2}||\boldsymbol{\theta}||^2,
\end{equation}
which is strongly convex and has Lipschitz gradient, and satisfies the conditions of Theorem \ref{th:error_bound_fading_channel}. In the simulations we set $\lambda=1/2$.

\subsubsection{Supporting the theoretical analysis under equal channel gains}
\begin{figure}[htbp]
\begin{center}
    \subfigure[Results for different (logspace scaling) values of $N$.] {\scalebox{0.4}
    {
      \label{fig:equal_channel_EN_1}{\epsfig{file=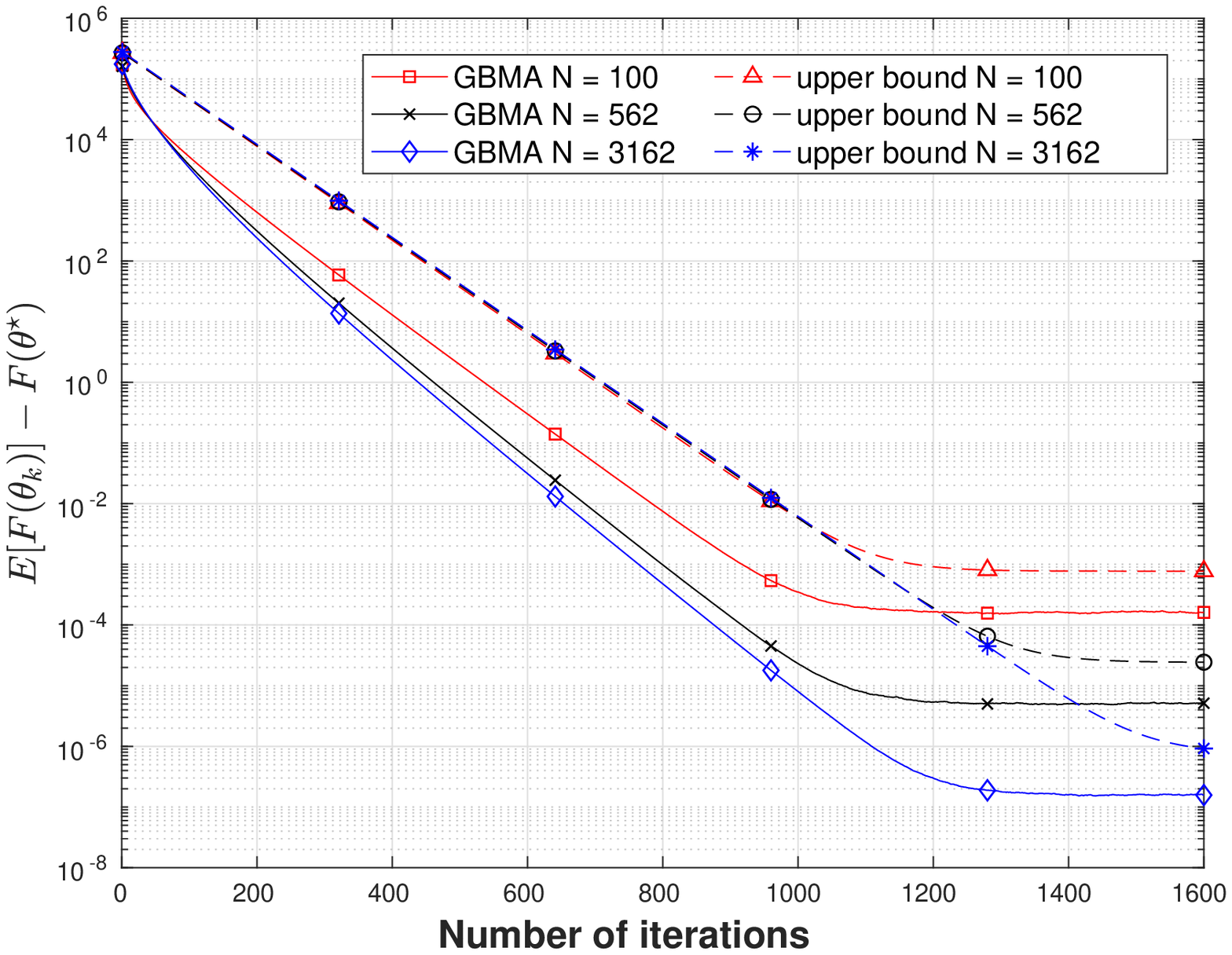,height=0.42\textheight,width=0.9\textwidth}}
    }}
    \subfigure[Results for $E_N=N^{\epsilon-2}$ using different values of $\epsilon$.]{\scalebox{0.4}
    {
      \label{fig:equal_channel_EN_epsilon}{\epsfig{file=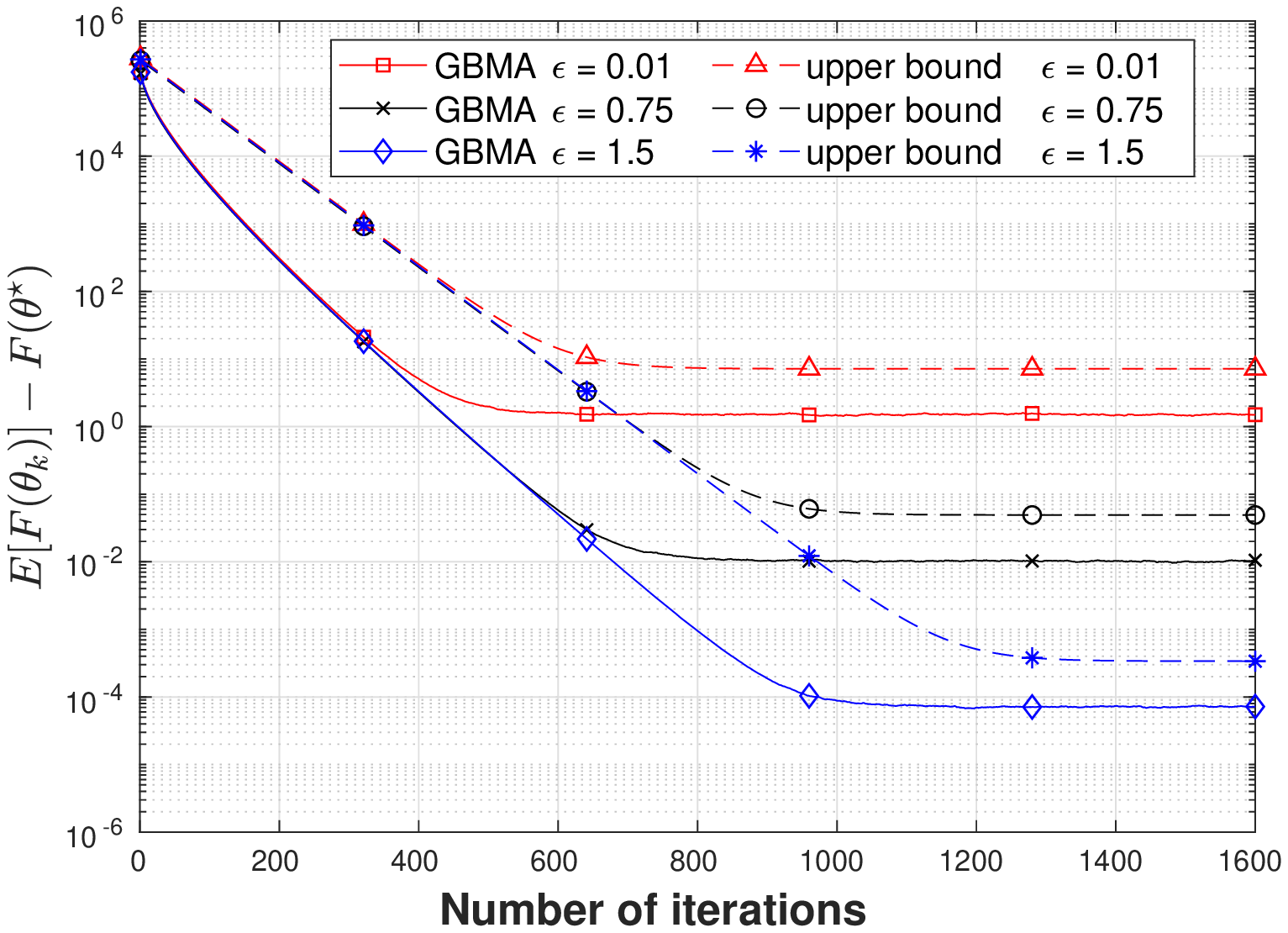,height=0.42\textheight,width=0.9\textwidth}}
    }}
   \caption{Simulation results for prediction of a release year of a song. The empirical errors and theoretical error bounds of GBMA algorithm are presented as a function of the number of iterations under equal channel gains.}
  \label{fig:equal_channel}
\end{center}
  \end{figure}

We start by examining the case of equal channel gains, i.e., $h_{n, k}=1$ for all $n, k$. The results are shown in Fig. \ref{fig:equal_channel}. 
In Fig. \ref{fig:equal_channel_EN_1}, we set the transmission energy to $E_N=1$, and used different values for $N$. In Fig. \ref{fig:equal_channel_EN_epsilon}, we set the number of nodes to $N=500$, and the transmission energy to $E_N = N ^{\epsilon - 2}$. We used different values for $\epsilon$. It can be seen that the empirical error supports the theoretical error bound, and has the same characterization as discussed in Section \ref{ssec:discussion}. The error decreases in a linear rate with the number of iterations for small $k$. This region is dominated by the initial distance error term. As $k$ increases, the variance term becomes dominant, and decreases by increasing $N$, or $E_N$.

\subsubsection{Supporting the theoretical analysis under Rayleigh fading channels}
 
\begin{figure}[htbp]
\begin{center}
    \subfigure[Results for different (logspace scaling) values of $N$.]{\scalebox{0.4}
    {
      \label{fig:fading_channel_EN_1}{\epsfig{file=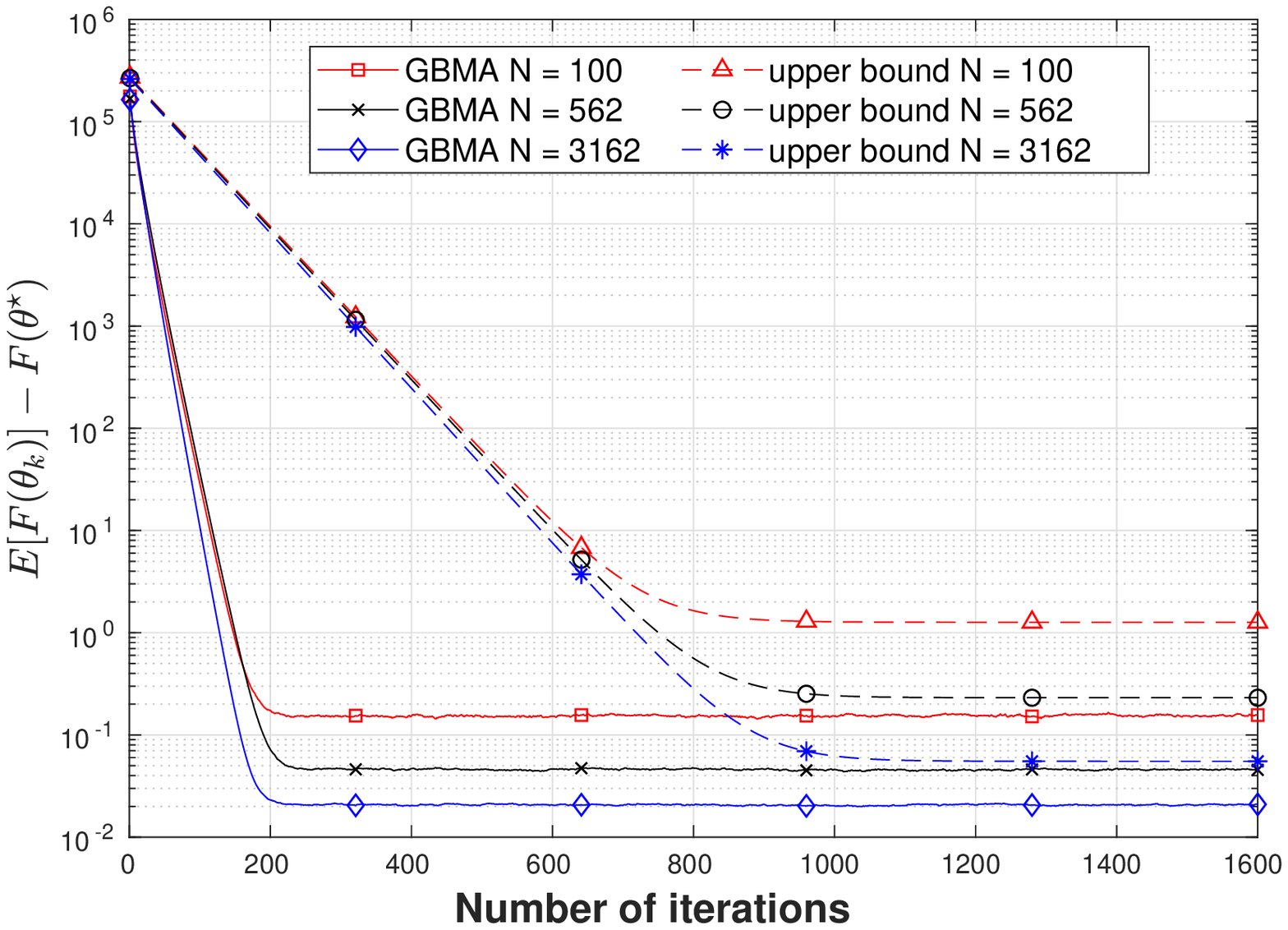,height=0.42\textheight,width=0.9\textwidth}}
    }}
    \subfigure[Results for $E_N=N^{\epsilon-2}$ using different values of $\epsilon$.]{\scalebox{0.4}
    {
      \label{fig:fading_channel_EN_epsilon}{\epsfig{file=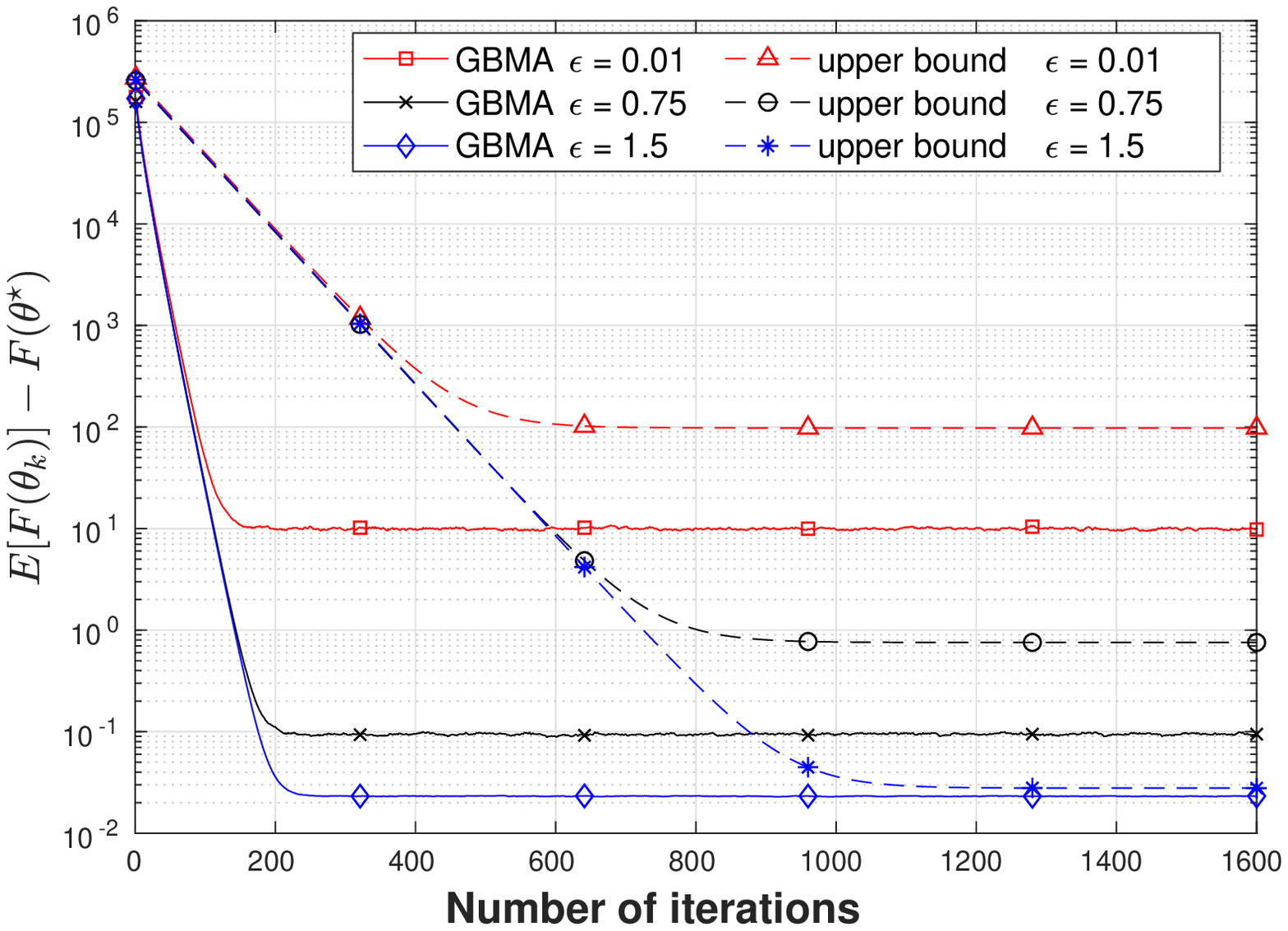,height=0.42\textheight,width=0.9\textwidth}}
    }}
   \caption{Simulation results for prediction of a release year of a song. The empirical errors and theoretical error bounds of GBMA algorithm are presented as a function of the number of iterations under i.i.d. Rayleigh fading channels.}
  \label{fig:fading_channel}
\end{center}
  \end{figure}

Next, we examine the case where the nodes experience i.i.d Rayleigh fading channel gains, $h_n \sim \text{Rayleigh}(\sigma_h)$. The results are shown in Fig. \ref{fig:fading_channel}. 
In Fig. \ref{fig:fading_channel_EN_1}, we set the transmission energy to $E_N=1$, and used different values for $N$. In Fig. \ref{fig:fading_channel_EN_epsilon}, we set the number of nodes to $N=500$, and the transmission energy to $E_N = N ^{\epsilon - 2}$. We used different values for $\epsilon$. It can be seen again that the empirical error supports the theoretical error bound, and has the same characterization as discussed in Section \ref{ssec:discussion}. The error decreases in a linear rate with the number of iterations for small $k$. This region is dominated by the initial distance error term. As $k$ increases, both the distortion term (due to the channel fading effect) and the variance term (due to the noisy channel) become dominant. The error decreases by increasing $N$, or $E_N$.

\subsubsection{Comparison with the centralized GD, and distributed GD using FDM}

Next, We compared GBMA with the following algorithms: (i) The centralized GD algorithm, in which the optimizer has access to each sample directly when updating the GD iterates. In this scheme there is no noise, nor channel fading effects in the system. It serves as a benchmark for comparison. (ii) The FDM-GD algorithm, in which each node is allocated a dedicated orthogonal channel for transmission. This scheme was widely used in federated learning applications (see e.g., \cite{chen2018lag, konevcny2016federated}). 

\begin{figure}[htbp]
\begin{center}{\scalebox{0.5}
  {\epsfig{file=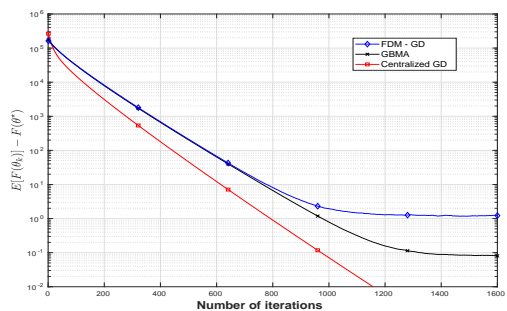,height=0.35\textheight,width=0.8\textwidth}}}
   \caption{Simulation results for prediction of a release year of a song. A comparison between GBMA, FDM-GD, and the centralized GD (which serves as a benchmark) is presented. GBMA was operated under $-50$dB, where FDM-GD was operated under $-6$dB.}
  \label{fig:fig_GBMAvsTDMA_Year}
\end{center}
  \end{figure}

The results are shown in Fig. \ref{fig:fig_GBMAvsTDMA_Year}. 
We examine the case where the nodes experience i.i.d Rayleigh fading channel gains. We set the number of nodes to $N=800$. Since the aggregated noise does not depend on $N$ under GBMA, we were able to use very low SNR as compared to FDM-GD and still outperforms it. Specifically, the energy coefficient was set to $E_N = N^{-1.5}$ under GBMA, and $E_N = 1$ under FDM-GD. As a result, GBMA was operated under $-50$dB, where FDM-GD was operated under $-6$dB. These results demonstrate the tremendous performance gain of GBMA over the traditional GD learning using orthogonal channels.

  \subsection{Source Localization using a Wireless Sensor Network}
\label{ssec:sourceLocalization}  

In this section we consider a parameter estimation problem of localizing a source that emits acoustic waves using wireless sensor networks, as in \cite{Rabbat2004localization,Rabbat2004DistOpt,Blatt_convergent_invcremental_const_step}. We simulated $N$ sensors which are located at known locations, denoted by $r_n \in \mathbb{R}^2, n=1,2,... ,N$. Each sensor holds a noisy measurement $x_n$ of the acoustic signal transmitted by the source at an unknown location $\boldsymbol{\theta} \in \mathbb{R}^2$. Based on the far field assumption, the measurement is modeled by $x_n = s_n + v_n$, where $v_n,  n=1,\cdots,N$ is an i.i.d Gaussian noise with zero mean, and 
  $$s_n = \frac{A}{||\boldsymbol{\theta} - r_n||^2},$$
  where $A$ is a known constant characterizing the signal strength of the source. The local loss function is given by:
  $$ f_n(\boldsymbol{\theta}) = \left( x_n - \frac{A}{||\boldsymbol{\theta} - r_n||^2}\right)^2,$$
which is not convex neither Liphschitz. As a result, the conditions of Theorems \ref{th:error_bound_fading_channel}, \ref{th:error_bound_fading_channel_convex} are not met. Nevertheless, when the source is sufficiently distant from the sensors, and the initial estimate is close to the true value, we demonstrate that GBMA succeeds to converge. 

\subsubsection{Comparison with the centralized GD, and distributed GD using FDM}
\begin{figure}[htp]
\begin{center}
    \subfigure[The error as a function of the number of iterations.]{\scalebox{0.4}
    {
      \label{fig:Inc_conv}{\epsfig{file=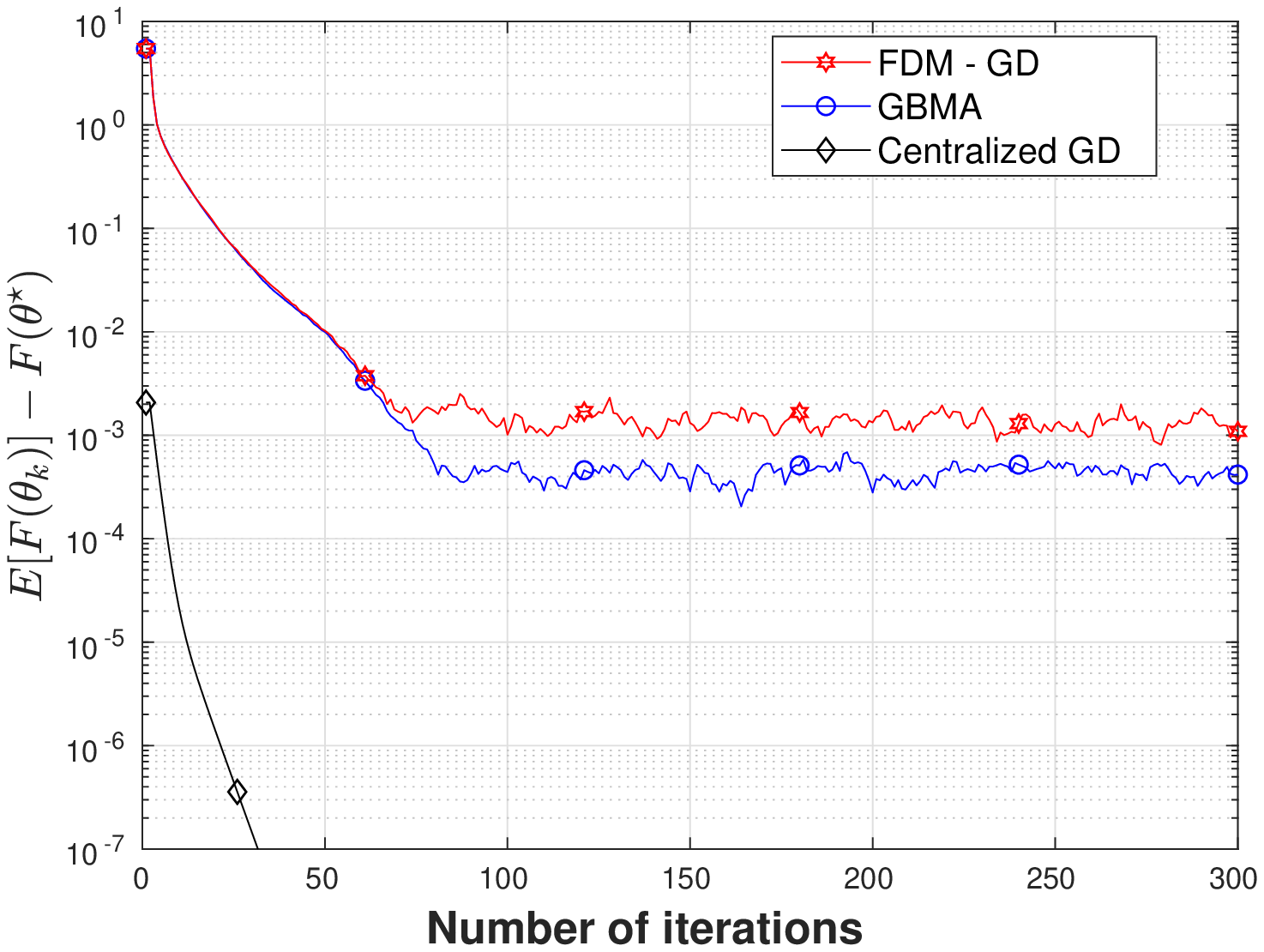,height=0.42\textheight,width=0.9\textwidth}}
    }}
    \subfigure[The total average transmission energy as a function of the error.]{\scalebox{0.4}
    {
      \label{fig:Inc_energy}{\epsfig{file=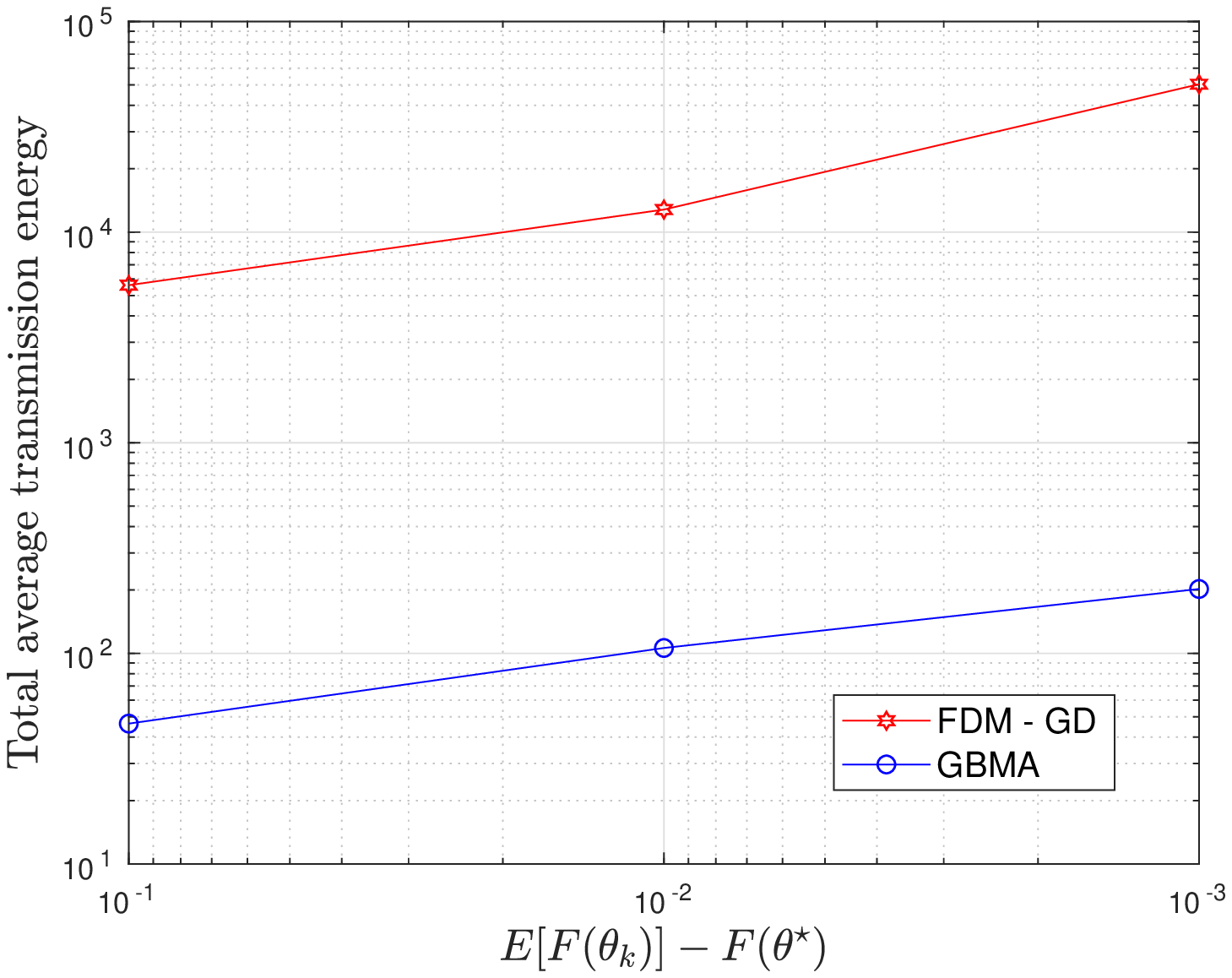,height=0.42\textheight,width=0.9\textwidth}}
    }}
   \caption{Simulation results for source localization. A comparison between GBMA, FDM-GD, and the centralized GD (which serves as a benchmark) is presented.}
  \label{fig:localization_comparison}
\end{center}
  \end{figure}

The number of sensors was set to $N = 200$, and they were randomly distributed on a perimeter of $100 X 100$m$^2$ field. Sensors were not located within a radius of $8$m from the source as they do not hold the far field assumption. The source was located at coordinates $[60,60]$. The received SNR was set to $-10$dB. The sampling noise was generated from a white normal Gaussian distribution. As can be seen in Fig. \ref{fig:localization_comparison}, the GBMA algorithm achieves a significant performance gain in terms of the empirical error, while saving a significant amount of transmission energy.
  
\subsubsection{Supporting the energy scaling laws}
\begin{figure}[htbp]
\begin{center}{\scalebox{0.35}
  {\epsfig{file=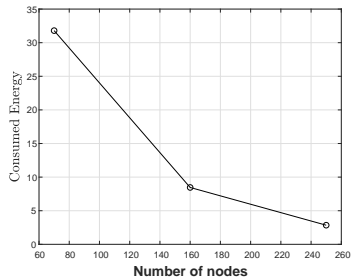}}}
   \caption{Simulation results for source localization. The total average transmission energy in the network under GBMA as a function of the number of sensors. The total energy decreases to zero, while keeping the error equals to $10^{-2}$.}
  \label{fig:localization_enregy}
\end{center}
  \end{figure}

The energy scaling laws discussed in Section \ref{ssec:discussion} state that we can reach any desired small error, while making the total transmission energy in the network arbitrarily close to zero by increasing the number of nodes and setting $N^{\epsilon-2}\lesssim E_N\lesssim N^{-\epsilon-1}$. We examined this result numerically by setting $E_N =N^{-1.5}$. Figure \ref{fig:localization_enregy} shows the decreasing of the total transmission energy to zero as $N$ increases, while keeping the desired error equals to $10^{-2}$. 

\section{Conclusion}
\label{sec:conclusion}
  
We considered a distributed learning problem over multiple access channel (MAC) using a large wireless network, where the objective function is a sum of the nodes' local loss functions. This problem has attracted a growing interest in distributed sensing systems, and more recently in federated learning. A novel Gradient-Based Multiple Access (GBMA) algorithm was developed and analyzed, in which the nodes transmit an analog function of the local gradient using a common shaping waveform and the network edge receives and update the estimate using a superposition of the analog transmitted signals which represents a noisy distorted version of the gradient. We established a finite-sample bound of the error for both convex and strongly convex loss functions with Lipschitz gradient. Furthermore, we provided specific energy scaling laws for approaching the centralized convergence rate as the number of nodes increases. Experimental results demonstrated strong performance of the GBMA algorithm and supported the theoretical results.

\section{Appendix}

In this appendix we provide the proofs for the main Theorems \ref{th:error_bound_fading_channel}, and \ref{th:error_bound_fading_channel_convex}.

\subsection{Proof of Theorem \ref{th:error_bound_fading_channel}}
\label{app:proof1}
Let 
\begin{equation}
\displaystyle r_k^2 \triangleq ||\boldsymbol{\theta_k} -\boldsymbol{\theta^*}||^2
\end{equation}
be the squared distance from $\boldsymbol{\theta^*}$ at time $k$. Recall that $\boldsymbol{\theta_{k+1}}=\boldsymbol{\theta_k}-\beta \boldsymbol{v_k}$. Thus, the squared distance at time $k+1$ can be written as: 
\begin{equation}
\begin{array}{c}
\displaystyle r_{k+1}^2 =||\boldsymbol{\theta_{k+1}} -\boldsymbol{\theta^*}||^2 \vspace{0.2cm}\\\hspace{1.7cm}
\displaystyle
=||(\boldsymbol{\theta_k} -\boldsymbol{\theta^*}) - \beta \boldsymbol{v_k}||^2.
\end{array}
\end{equation}
By replacing $r_k^2 = ||\boldsymbol{\theta_k} -\boldsymbol{\theta^*}||^2$, we can write:
\begin{equation}
\label{eq:r_k_raw_fading_channel}
 r_{k+1} ^2 = r_k^2 - 2\beta  \boldsymbol{v_k}^T\left(\boldsymbol{\theta_k}-\boldsymbol{\theta^*}\right) +\beta^2 ||\boldsymbol{v_k}||^2
\end{equation}

Before  proceeding, we first evaluate the expected values of $\boldsymbol{v_k}$, and $||\boldsymbol{v_k}||^2$ with respect to the additive noise and channel gain processes up to time $k$.
The expected value $\mathbb{E}[\boldsymbol{v_k}]$ of $\boldsymbol{v_k}$ is given by:
\begin{equation}
\label{eq:mean_of_v_k_fading_channel}
\begin{array}{l}
\displaystyle \mathbb{E}[\boldsymbol{v_k}] = \mathbb{E}\left [\frac{1}{N} \sum_{n=1}^N h_{n,k}\boldsymbol{g_n}(\boldsymbol{\boldsymbol{\theta_k}}) +\boldsymbol{w_k}\right ]  
\vspace{0.2cm}\\\hspace{0.9cm}
=\mu_h\mathbb{E}\left[\boldsymbol{\nabla} F(\boldsymbol{\theta_k})\right]+\mathbb{E}[\boldsymbol{w_k}]
\vspace{0.2cm}\\\hspace{0.9cm}
=\mu_h\mathbb{E}\left[\boldsymbol{\nabla} F(\boldsymbol{\theta_k})\right],
\end{array}
\end{equation}
where the second equality follows since $h_{n,k}$ is independent of $\boldsymbol{\theta_k}$.

The expected value $\mathbb{E}\left[||\boldsymbol{v_k}||^2\right]$ of $||\boldsymbol{v_k}||^2$ is given by:
\begin{equation}
\label{eq:mean_of_||v_k||^2_fading_channel}
\begin{array}{l}
\displaystyle\mathbb{E}\left[|| \boldsymbol{v_k} ||^2\right]
 = \mathbb{E}\left[\left|\left|\frac{1}{N} \sum_{n=1}^N h_{n,k}\boldsymbol{g_n}(\boldsymbol{\theta_k})+\boldsymbol{w_k}\right|\right|^2 \right] 
 \vspace{0.2cm}\\\hspace{1.6cm}
\displaystyle
= \mathbb{E}\left[\left|\left|\frac{1}{N} \sum_{n=1}^N h_{n,k}\boldsymbol{g_n}(\boldsymbol{\theta_k})\right|\right|^2 \right]
\vspace{0.2cm}\\\hspace{2cm}
\displaystyle
+2\mathbb{E}\left[\left(\frac{1}{N} \sum_{n=1}^N h_{n,k}\boldsymbol{g_n}(\boldsymbol{\theta_k})\right)^T \boldsymbol{w_k}\right] 
\vspace{0.2cm}\\\hspace{2cm}
\displaystyle
+\mathbb{E}\left[\left|\left|\boldsymbol{w_k}\right|\right|^2 \right]
\vspace{0.2cm}\\\hspace{1.3cm}
\displaystyle
= \frac{1}{N^2} \sum_{n,m =1}^N \mathbb{E}\left[(h_{n,k} \boldsymbol{g_n}(\boldsymbol{\theta_k}))^T (h_{m,k}  \boldsymbol{g_m}(\boldsymbol{\theta_k}))\right] 
\vspace{0.2cm}\\\hspace{2cm}
\displaystyle
+\frac{ d\sigma_w^2} {E_N N^2}.
\end{array}
\end{equation}

The term $\mathbb{E}\left[(h_{n,k} \boldsymbol{g_n}(\boldsymbol{\theta_k}))^T (h_{m,k}  \boldsymbol{g_m}(\boldsymbol{\theta_k}))\right]$ can be written as:
\begin{equation}
\label{eq:h_g^T_h_g}
   \begin{array}{l}  
 \mathbb{E}[(h_{n,k}  \boldsymbol{g_n}(\boldsymbol{\theta_k})^T (h_{m,k}  \boldsymbol{g_m}(\boldsymbol{\theta_k})]     \vspace{0.2cm}\\ \hspace{0.4cm} \displaystyle 
=\mathbb{E}[h_{n,k}h_{m,k}] \mathbb{E}[\boldsymbol{g_n}(\boldsymbol{\theta_k})^T \boldsymbol{g_m}(\boldsymbol{\theta_k})] \vspace{0.2cm}\\ \hspace{0.4cm} \displaystyle 
 =(\mu_h^2 +\sigma_h^2 \textbf{1}_{\left\{n=m\right\}})\mathbb{E}[\boldsymbol{g_n}(\boldsymbol{\theta_k})^T \boldsymbol{g_m}(\boldsymbol{\theta_k})],
 \end{array} 
\end{equation}
where $\textbf{1}_{\left\{n=m\right\}}=1$ if $n=m$, and $\textbf{1}_{\left\{n=m\right\}}=0$ otherwise. The second equality follows since $h_{n,k}, h_{m,k}$ are independent of $\boldsymbol{g_n}(\boldsymbol{\theta_k}), \boldsymbol{g_m}(\boldsymbol{\theta_k})$ for all $n, m$. The last equality follows since $h_{n,k}, h_{m,k}$ are independent between them for $n\neq m$.
\noindent
Substituting (\ref{eq:h_g^T_h_g}) in \eqref{eq:mean_of_||v_k||^2_fading_channel} yields:
\begin{equation}
\label{eq:mean_of_||v_k||^2_fading_channel2}
\begin{array}{l}
\displaystyle\mathbb{E}\left[|| \boldsymbol{v_k} ||^2\right]=
\vspace{0.2cm}\\\hspace{0.5cm}
\displaystyle
\mu_h^2 \mathbb{E}\left[||\boldsymbol{\nabla} F(\boldsymbol{\theta_k})||^2\right] + \frac{\sigma_h^2}{N^2}\sum_{n=1}^N  \mathbb{E}\left[||\boldsymbol{\nabla} f_n(\boldsymbol{\theta_k})||^2\right] 
\vspace{0.2cm}\\\hspace{2cm}
\displaystyle
+\frac{ d\sigma_w^2} {E_N N^2}.
\end{array}
\end{equation}
\noindent

Next, we take expectation of both parts of \eqref{eq:r_k_raw_fading_channel} to get:

\begin{equation} 
\label{eq:r_k mean_fading_channel}
\begin{array}{l}
\displaystyle\mathbb{E}[r_{k+1}^2] 
= \mathbb{E}[r_k^2] -2\beta \mathbb{E}\left[\boldsymbol{v_k}^T\left( \boldsymbol{\theta_k}-\boldsymbol{\theta^*} \right)\right] 
\displaystyle+ \beta^2 \mathbb{E}\left[||\boldsymbol{v_k}||^2\right]
\vspace{0.2cm}\\\hspace{1.23cm} 
\displaystyle
= \mathbb{E}[r_k^2] -2\beta\mu_h \mathbb{E}\left[ \boldsymbol{\nabla} F(\boldsymbol{\theta_k})^T\left( \boldsymbol{\theta_k}-\boldsymbol{\theta^*} \right)\right] \vspace{0.2cm}\\\hspace{1.4cm} 
\displaystyle+ \beta^2\mu_h^2 \mathbb{E}\left[||\boldsymbol{\nabla} F(\boldsymbol{\theta_k})||^2\right]\vspace{0.2cm}\\\hspace{1.4cm} 
\displaystyle+  \frac{\beta^2\sigma_h^2}{N^2}\sum_{n=1}^N  \mathbb{E}\left[||\boldsymbol{\nabla} f_n(\boldsymbol{\theta_k})||^2\right] 
+\beta^2\frac{ d\sigma_w^2} {E_N N^2},
\end{array}
\end{equation}
where the second term on the RHS of (\ref{eq:r_k mean_fading_channel}) holds since $\boldsymbol{w_k}$ is independent of $\boldsymbol{\theta_k}$, and the third, fourth, and fifth terms follows by (\ref{eq:mean_of_||v_k||^2_fading_channel2}).

We next exploit the $L$-Lipschitz gradient and strongly convex properties of $F(\boldsymbol{\theta})$. Taking expectation of \eqref{eq: 2.1.15 inequality} and setting $\boldsymbol{\nabla} F(\boldsymbol{\theta^*}) =0$ yields:
\begin{equation} 
\label{eq:2.1.15_implementation}
\begin{array}{l}
\displaystyle\mathbb{E}\left[\boldsymbol{\nabla} F(\boldsymbol{\theta_k})^T \left(\boldsymbol{\theta_k}-\boldsymbol{\theta^*} \right)\right]
\vspace{0.2cm}\\\hspace{0cm}
\displaystyle\geq 
\frac{\mu L}{\mu+L} \mathbb{E}[r_k^2] + \frac{1}{\mu +L}\mathbb{E}\left[||\boldsymbol{\nabla} F(\boldsymbol{\theta_k})||^2\right].
\end{array}
\end{equation}
By substituting (\ref{eq:2.1.15_implementation}) in \eqref{eq:r_k mean_fading_channel} we get
\begin{equation}
\label{eq:r_k_1_after_lips_fading_channel}
\begin{array}{l}
\displaystyle\mathbb{E}[r_{k+1}^2] \vspace{0.2cm}\\\hspace{0cm}
\displaystyle
\leq \mathbb{E}[r_k^2] -2\beta\mu_h \left(\frac{\mu L}{\mu+L} \mathbb{E}[r_k^2] + \frac{1}{\mu +L}\mathbb{E}\left[||\boldsymbol{\nabla} F(\boldsymbol{\theta_k})||^2\right]\right) \vspace{0.2cm}\\\hspace{1.4cm} 
\displaystyle+ \beta^2\mu_h^2 \mathbb{E}\left[||\boldsymbol{\nabla} F(\boldsymbol{\theta_k})||^2\right]\vspace{0.2cm}\\\hspace{1.4cm} 
\displaystyle+  \frac{\beta^2\sigma_h^2}{N^2}\sum_{n=1}^N  \mathbb{E}\left[||\boldsymbol{\nabla} f_n(\boldsymbol{\theta_k})||^2\right] 
+\beta^2\frac{ d\sigma_w^2} {E_N N^2}.
\end{array}
\end{equation}

Let 
\begin{equation}\label{tilde_sigma_strong_convex_fading}
\displaystyle\tilde{\sigma}\triangleq\frac{\sigma_h^2}{N^2}\sum_{n=1}^N  \mathbb{E}\left[||\boldsymbol{\nabla} f_n(\boldsymbol{\theta_k})||^2\right]+\frac{d\sigma_w^2} {E_N N^2}.
\end{equation}

Let $\theta^*_n$ be the minimizer of $f_n$. Since $f_n$ is $L_n$-Lipschitz, we have:
\begin{equation}
\begin{array}{l}
     \displaystyle \mathbb{E}\left[ ||\nabla f_n(\tk)||^2 \right] \leq L_n^2 \mathbb{E}\left[||\tk- \boldsymbol{\theta^{\star}_n}||^2\right]  \vspace{0.3cm}\\ \hspace{0.2cm}
\displaystyle
= L_n^2\mathbb{E}[ ||\tk -\tstar +\tstar -\boldsymbol{\theta^{\star}_n}||^2]  
\vspace{0.3cm}\\ \hspace{0.2cm}
\displaystyle
= L_n^2\left( \mathbb{E}[ ||\tk -\tstar ||^2]\right.  \vspace{0.2cm}\\ \hspace{0.5cm}
\displaystyle \left. + 2\mathbb{E}[\left(\tk -\tstar\right)^T\left(\tstar -\boldsymbol{\theta^{\star}_n}  \right) ] +\mathbb{E}[ ||\tstar -\boldsymbol{\theta^{\star}_n}||^2] \right) \vspace{0.2cm}\\ \hspace{0.3cm}
\displaystyle 
\leq L_n^2 \mathbb{E}[r_k^2]+ 2L_n^2 \mathbb{E}[r_k \cdot  ||\tstar -\boldsymbol{\theta^{\star}_n }||] +L_n^2 \delta^2  \vspace{0.2cm}\\ \hspace{0.3cm}
\displaystyle 
\leq \overline{L}^2 \mathbb{E}[r_k^2]+ 2\overline{L}^2\delta\mathbb{E}[r_k] + \overline{L}^2\delta^2
\end{array}
\end{equation}
where the first inequality holds by the Cauchy-Schwartz inequality.
Then, the summation in (\ref{tilde_sigma_strong_convex_fading}) can be bounded by:
\begin{equation} 
\label{eq: bound_sum_of_gradients}
\begin{array}{l}
    \displaystyle
    \frac{1}{N^2}\sum_{n=1}^N  \mathbb{E}\left[||\boldsymbol{\nabla} f_n(\tk)||^2\right]
    \vspace{0.2cm}\\\hspace{1cm}
    \displaystyle
    \leq\frac{\overline{L}^2}{N} \mathbb{E}[r_k^2]+\frac{2\overline{L}^2 \delta}{N} \mathbb{E}[r_k] +\frac{\overline{L}^2 \delta^2}{N}.
\end{array}
\end{equation}
Using this result, we can upper bound $\tilde{\sigma}$:
\begin{equation}
  \displaystyle \tilde{\sigma} \leq  \frac{\sigma_h^2\overline{L}^2}{N} \mathbb{E}[r_k^2]+  \frac{2\sigma_h^2\overline{L}^2 \delta}{N} \mathbb{E}[r_k] + \frac{\sigma_h^2\overline{L}^2 \delta^2}{N}  + \frac{d\sigma_w^2} {E_N N^2}.
\end{equation}

Then, we can rewrite (\ref{eq:r_k_1_after_lips_fading_channel}) as: 
\begin{equation}
\begin{array}{l}\label{eq:r_k+1^2 step2 }
\displaystyle\mathbb{E}[r_{k+1}^2] \leq \tilde{c} \mathbb{E}[r_k^2]+  b  \mathbb{E}[r_k]
\vspace{0.2cm}\\\hspace{0.5cm}
\displaystyle
+\beta\left(\beta\mu_h^2-\frac{2\mu_h}{\mu+L}\right)
\mathbb{E}\left[||\nabla F(\tk)||^2\right] \vspace{0.2cm}\\\hspace{0.5cm}
\displaystyle + \beta^2\left( \frac{\sigma_h^2 
\overline{L}^2 \delta^2}{N} +\frac{d\sigma_w^2} {E_N N^2}\right),
\end{array}
\end{equation}
where
\begin{equation}
\begin{array}{l}
\displaystyle
\tilde{c}\triangleq 1-\frac{2\beta\mu_h\mu L}{\mu +L} + \beta^2 \sigma_h^2 \frac{\overline{L}^2}{N},\vspace{0.2cm}\\\displaystyle
b\triangleq  \frac{2\beta^2 \sigma_h^2\overline{L}^2 \delta}{N}.
\end{array}
\end{equation}
We also define:
\begin{equation}
\begin{array}{l}
\displaystyle
c\triangleq \tilde{c}+b.
\end{array}
\end{equation}
Notice that 
\begin{equation}
    \tilde{c}\mathbb{E}[r_k^2] +b\mathbb{E}[r_k] \leq c\mathbb{E}[r_k^2] +b
\end{equation}
In addition, note that condition \eqref{eq:step_condition_strong_convex_gradient_unbounded} implies that $0<c<1$. 
Also, condition \eqref{eq:step_condition_strong_convex_gradient_unbounded} implies
\begin{center}
$\displaystyle \beta\left(\beta\mu_h^2-\frac{2\mu_h}{\mu+L}\right)
\mathbb{E}\left[||\boldsymbol{\nabla} F(\boldsymbol{\theta_k})||^2\right]<0$.   
\end{center}
As a result, 
\begin{equation}
\displaystyle
\mathbb{E}[r_{k+1}^2] \leq c E[r_{k}^2]+\beta^2\left(\frac{\sigma_h^2 \delta\overline{L}^2(2+\delta)}{N} +\frac{d\sigma_w^2} {E_N N^2}\right).
\end{equation}
Similarly,  
\begin{equation}
\displaystyle
\mathbb{E}[r_{k}^2] \leq c E[r_{k-1}^2]+\beta^2\left(\frac{\sigma_h^2 \delta\overline{L}^2(2+\delta)}{N} +\frac{d\sigma_w^2} {E_N N^2}\right).
\end{equation}
Combining the last two inequalities yields: 
\begin{equation}
\begin{array}{l}
\displaystyle \mathbb{E}[r_{k+1}^2] \leq c \left(c E[r_{k-1}^2]+\beta^2\left(\frac{\sigma_h^2 \delta\overline{L}^2(2+\delta)}{N} +\frac{d\sigma_w^2} {E_N N^2}\right)\right)
\vspace{0.2cm}\\\hspace{2cm}\displaystyle
+\beta^2\left(\frac{\sigma_h^2 \delta\overline{L}^2(2+\delta)}{N} +\frac{d\sigma_w^2} {E_N N^2}\right),
\end{array}
\end{equation}
and so by induction we reach
\begin{equation}
\label{eq: r_k almost done_equal_channel}
\displaystyle
\mathbb{E}[r_{k+1}^2] \leq c^k r_0^2 +\beta^2\left(\frac{\sigma_h^2 \delta\overline{L}^2(2+\delta)}{N} +\frac{d\sigma_w^2} {E_N N^2}\right) \sum_{i=0}^{k-1} c^i.
\end{equation}
Calculating the bound of the sum geometric series yields:
\begin{equation}
\label{eq:sum_geometric_equal_channel}
\displaystyle \sum_{i=0}^{k-1} c^i \leq  \sum_{i=0}^{\infty} c^i = \frac{1}{1-c}.
\end{equation}

Substituting (\ref{eq:sum_geometric_equal_channel}) in (\ref{eq: r_k almost done_equal_channel}) yields:
\begin{equation}
\label{eq: r_k almost done_fading_channel_not_bounded_gradients}
\displaystyle
\mathbb{E}[r_{k+1}^2] \leq c^k r_0^2 +\frac{\beta^2}{1-c}\left(\frac{\sigma_h^2 \delta\overline{L}^2(2+\delta)}{N} +\frac{d\sigma_w^2} {E_N N^2}\right).
\end{equation}
Finally, applying lemma \ref{lemma:lip_ineuality}, setting $\boldsymbol{\nabla} F(\tstar) = 0$, and taking expectation yield: 
\begin{multline}
\mathbb{E}[F(\tk)] - F(\tstar) \leq \frac{L}{2} \mathbb{E}\left[||\tk - \tstar||^2\right] \\ \leq c^k r_0^2\frac{L}{2} +\frac{L\beta^2}{2(1-c)}\left(\frac{\sigma_h^2 \delta\overline{L}^2(2+\delta)}{N} +\frac{d\sigma_w^2} {E_N N^2}\right),
\end{multline}
which completes the proof. 
\hfill $\square$

\subsection{Proof of Theorem \ref{th:error_bound_fading_channel_convex}}
\label{app:proof2}  

We start by proving part a. 
By applying Lemma \ref{lemma:lip_ineuality} and taking expectation we get
\begin{equation}  
\label{eq:start of convex proof_equal}
\begin{array}{l}
\displaystyle
\mathbb{E}[F(\boldsymbol{\theta_{k+1}}]   \leq  \mathbb{E}[F(\boldsymbol{\theta_k})] + \mathbb{E}\left[ \boldsymbol{\nabla} F(\boldsymbol{\theta_k})^T \left(\boldsymbol{\theta_{k+1}} -\boldsymbol{\theta_k}\right) \right] \vspace{0.2cm} \\ \hspace{4cm} 
\displaystyle
+\frac{L}{2}\mathbb{E}\left[|| \boldsymbol{\theta_{k+1}} -\boldsymbol{\theta_k} ||^2\right]. 
\end{array} 
\end{equation}
From \eqref{eq:GD}, we have $\boldsymbol{\theta_{k+1}} -\boldsymbol{\theta_k} = -\beta \boldsymbol{v_k}$, and by a applying \eqref{eq:v_k definition} and \eqref{eq:mean_of_||v_k||^2_fading_channel2} with $h_{n,k}=1, \sigma_h^2=0$ we can write the term
$\mathbb{E}\left[ \left<\boldsymbol{\nabla} F(\boldsymbol{\theta_k}), \boldsymbol{\theta_{k+1}} -\boldsymbol{\theta_k}\right > \right]$ as
\begin{equation}
\begin{array}{l}
\displaystyle
\mathbb{E}\left[ \boldsymbol{\nabla} F(\boldsymbol{\theta_k})^T \left(-\beta \boldsymbol{v_k} \right) \right] = -\beta  \mathbb{E}\left[ \boldsymbol{\nabla} F(\boldsymbol{\theta_k})^T\left(  \boldsymbol{\nabla} F(\boldsymbol{\theta_k}) + \boldsymbol{w_k} \right) \right] 
\vspace{0.2cm}\\ \hspace{1cm} 
\displaystyle
= -\beta \mathbb{E}\left[ || \boldsymbol{\nabla} F(\boldsymbol{\theta_k})||^2\right]
\vspace{0.2cm}\\ \hspace{1.1cm}
\displaystyle
 = -\beta \mathbb{E}\left[ ||\boldsymbol{v_k}||^2 \right] +\beta \frac{d\sigma_w^2}{E_N N^2}
\end{array} 
\end{equation}
Inserting this result back to \eqref{eq:start of convex proof_equal} yields:
\begin{equation}
\begin{array}{l}
\displaystyle\mathbb{E}[F(\boldsymbol{\theta_{k+1}}]   
\vspace{0.2cm}\\\hspace{0.5cm}\displaystyle
\leq  \mathbb{E}[F(\boldsymbol{\theta_k})] -\beta\left( 1 - \frac{L}{2}\beta\right)\mathbb{E}\left[ ||\boldsymbol{v_k}||^2 \right] +\beta \frac{d\sigma_w^2}{E_N N^2}. 
\end{array}
\end{equation}
By condition \eqref{eq: beta condition_equal_convex} we get that $-\left( 1 - \frac{L}{2}\beta\right) \leq -\frac{1}{2}$. Hence, 
\begin{equation} \label{eq:F_k+1 bound_equal}
\mathbb{E}[F(\boldsymbol{\theta_{k+1}}]   \leq  \mathbb{E}[F(\boldsymbol{\theta_k})] -\frac{\beta}{2}\mathbb{E}\left[ ||\boldsymbol{v_k}||^2 \right] +\beta \frac{d\sigma_w^2}{E_N N^2}. 
\end{equation}
The sequence \{$ \mathbb{E}[F(\boldsymbol{\theta}_i)]$\} is monotonically decreasing for $i=1, ..., k+1$ if the following condition holds: 
\begin{equation} \label{eq:condition for decreasing F}
\begin{array}{l}
\displaystyle\mathbb{E}\left[ ||v_i||^2 \right] = \mathbb{E}\left[ || \boldsymbol{\nabla} F(\boldsymbol{\theta}_i)||^2\right] + \frac{d\sigma_w^2}{E_N N^2}  \geq  \frac{2d\sigma_w^2}{E_N N^2},
\end{array}
\end{equation}
which can be written as
\begin{equation}
\begin{array}{l}
\displaystyle
\mathbb{E}\left[ || \boldsymbol{\nabla} F(\boldsymbol{\theta_k})||^2\right] >  \frac{d\sigma_w^2}{E_N N^2},\;\;\mbox{for}\;\; i=1, ..., k,
\end{array}
\end{equation}
which is satisfied by condition (\ref{eq:condition_decreasing}).
Next, by the convexity of $F(\boldsymbol{\theta})$ we get 
\begin{equation}
\mathbb{E}\left[F(\boldsymbol{\theta_k})\right] \leq F(\boldsymbol{\theta^*}) + \mathbb{E}\left[ \boldsymbol{\nabla} F(\boldsymbol{\theta_k})^T\left( \boldsymbol{\theta}_{k} -\boldsymbol{\theta^*}\right) \right].
\end{equation}
Substituting the last inequality into \eqref{eq:F_k+1 bound_equal} yields:
\begin{equation} 
\begin{array}{l}
\mathbb{E}\left[F(\boldsymbol{\theta_{k+1}})\right] \leq F(\boldsymbol{\theta^*}) +\mathbb{E}\left[ \boldsymbol{\nabla} F(\boldsymbol{\theta_k})^T\left( \boldsymbol{\theta}_{k} -\boldsymbol{\theta^*}\right) \right] 
\vspace{0.3cm}\\ \hspace{2cm}
\displaystyle -\frac{\beta}{2}\mathbb{E}\left[ ||\boldsymbol{v_k}||^2 \right] +\beta \frac{d\sigma_w^2}{E_N N^2}.
\end{array}
\end{equation}
As a result,
\begin{equation} 
\begin{array}{l}
\displaystyle
\mathbb{E}\left[F(\boldsymbol{\theta_{k+1}})- F(\boldsymbol{\theta^*}) \right] \leq \frac{1}{2\beta}\left(2\beta \mathbb{E}\left[ \left(\boldsymbol{v_k}-\boldsymbol{w_k}\right)^T\left(\boldsymbol{\theta}_{k} -\boldsymbol{\theta^*}\right) \right] \right.
\vspace{0.3cm}\\ \hspace{4cm} \left.
\displaystyle  -\beta^2 \mathbb{E}\left[ ||\boldsymbol{v_k}||^2 \right]  \right) +  \beta \displaystyle\frac{d\sigma_w^2}{E_N N^2} 
\vspace{0.2cm}\\ \hspace{0.2cm}
\displaystyle
= \frac{1}{2\beta}\left( 2\beta \mathbb{E}\left[ \boldsymbol{v_k}^T\left( \boldsymbol{\theta}_{k} -\boldsymbol{\theta^*}\right) \right] -\beta^2 \mathbb{E}\left[ ||\boldsymbol{v_k}||^2 \right]  \right.
\vspace{0.3cm}\\ \hspace{1cm}\left.
\displaystyle  -\mathbb{E}[||\boldsymbol{\theta_k}-\boldsymbol{\theta^*}||^2]+ \mathbb{E}[||\boldsymbol{\theta_k}-\boldsymbol{\theta^*}||^2] \right) +  \beta \displaystyle\frac{d\sigma_w^2}{E_N N^2} 
\vspace{0.2cm}\\ \hspace{0.2cm}
\displaystyle
= \frac{1}{2\beta}\left(- \mathbb{E}[||\boldsymbol{\theta_k} -\beta \boldsymbol{v_k} -\boldsymbol{\theta^*}||^2] + \mathbb{E}||\boldsymbol{\theta_k}-\boldsymbol{\theta^*}||^2]\right) 
\vspace{0.2cm}\\ \hspace{6cm}
\displaystyle
+  \beta \frac{d\sigma_w^2}{E_N N^2} 
\vspace{0.2cm}\\ \hspace{0.2cm}
\displaystyle
= \frac{1}{2\beta}\left(\mathbb{E}[r_k^2] - \mathbb{E}[r_{k+1}^2]\right) + \beta \frac{d\sigma_w^2}{E_N N^2}.
\end{array}
\end{equation}
Since the sequence \{$ \mathbb{E}[F(\boldsymbol{\theta}_i)]$\} is monotonically decreasing for $i=1, ..., k+1$, we can write: 
\begin{equation}
\label{eq:F_tk upper bound}
\begin{array}{l}
\displaystyle\mathbb{E}\left[F(\boldsymbol{\theta}_{k})- F(\boldsymbol{\theta^*}) \right] \leq \frac{1}{k} \sum_{i=1}^k  \mathbb{E}\left[F(\boldsymbol{\theta}_{i})- F(\boldsymbol{\theta^*}) \right] \vspace{0.3cm}\\ \hspace{0.2cm}
\displaystyle
\leq  \frac{1}{k} \sum_{i=1}^k \left( \frac{1}{2\beta}\left(\mathbb{E}[r_{i-1}^2] - \mathbb{E}[r_{i}^2]\right) + \beta \frac{d\sigma_w^2}{E_N N^2} \right) \vspace{0.3cm}\\ \hspace{0.2cm}
\displaystyle
\leq \frac{1}{2\beta k}(r_0^2 - \mathbb{E}[r_{k}^2]) +\beta \frac{d\sigma_w^2}{ E_N N^2}
\vspace{0.3cm}\\ \hspace{0.2cm}
\displaystyle
\leq \frac{1}{2\beta k}r_0^2 + \beta \frac{d\sigma_w^2}{ E_N N^2},
\end{array}
\end{equation}
where the third inequality holds due to the telescoping sum. This completes the proof of part a. 

Next, we prove part b. 
We start by bounding $\mathbb{E}[F(\boldsymbol{\theta_{k+1}}]$ for the fading channel case using the inequality that we obtained in (\ref{eq:start of convex proof_equal}). From \eqref{eq:GD} we have $\boldsymbol{\theta_{k+1}} -\boldsymbol{\theta_k} = -\beta \boldsymbol{v_k}$, and by applying \eqref{eq:v_k definition}, we get

\begin{equation} 
\label{eq: F_k+1_bound_FadingChannel}
    \begin{array}{l}
\mathbb{E}[F(\boldsymbol{\theta_{k+1}}] \leq  \mathbb{E}[F(\boldsymbol{\theta_k})] -\beta \mathbb{E}\left[ \boldsymbol{\nabla} F(\boldsymbol{\theta_k})^T\boldsymbol{v_k} \right] 
\vspace{0.2cm} \\ \hspace{5cm} 
\displaystyle
+\frac{L}{2} \beta^2\mathbb{E} \left[|| \boldsymbol{v_k} ||^2\right]
\vspace{0.2cm} \\ \hspace{0.0cm} 
\displaystyle
=   \mathbb{E}[F(\boldsymbol{\theta_k})] 
\vspace{0.2cm} \\ \hspace{0.0cm} 
\displaystyle
-\beta \mathbb{E}\left[ \boldsymbol{\nabla} F(\boldsymbol{\theta_k})^T\left( \frac{1}{N} \sum_{n=1}^{N}  h_{n,k} \boldsymbol{g_n}(\boldsymbol{\theta_k}) + \boldsymbol{w_k} \right) \right]
  \vspace{0.2cm} \\ \hspace{0.0cm} 
\displaystyle
+ \frac{L}{2}\beta^2\left( \mu_h^2 \mathbb{E}\left[||\boldsymbol{\nabla} F(\boldsymbol{\theta_k})||^2\right] + \frac{\sigma_h^2}{N^2}\sum_{n=1}^N  \mathbb{E}\left[||\boldsymbol{\nabla} f_n(\boldsymbol{\theta_k})||^2\right] \right.
\vspace{0.2cm}\\ \hspace{6cm}
\displaystyle
\left. +\frac{ d\sigma_w^2} {E_N N^2}  \right),
    \end{array}
\end{equation}
where $\mathbb{E}\left[||\boldsymbol{v_k}||^2\right]$ was computed in (\ref{eq:mean_of_||v_k||^2_fading_channel2}).
Let
\begin{equation}
    \tilde{\sigma}_k \triangleq  \frac{\sigma_h^2}{N^2}\sum_{n=1}^N  \mathbb{E}\left[||\boldsymbol{\nabla} f_n(\boldsymbol{\theta_k})||^2\right]+\frac{ d\sigma_w^2} {E_N N^2}.
\end{equation}
Substituting $\tilde{\sigma}_k$ into \eqref{eq: F_k+1_bound_FadingChannel}, and using the fact that $\boldsymbol{w_k}, \boldsymbol{\theta_k}$, and $h_{n,k}$ are independent, yield:
\begin{equation}
    \begin{array}{l}
\displaystyle  \mathbb{E}[F(\boldsymbol{\theta_{k+1}}] \leq  \mathbb{E}[F(\boldsymbol{\theta_k})]
\vspace{0.2cm}\\ \hspace{0.2cm}
\displaystyle
-\beta \mu_h \left(1-\frac{L\beta}{2}\mu_h\right)\mathbb{E}[||\boldsymbol{\nabla} F(\boldsymbol{\theta_k})||^2] 
+\frac{L\beta^2}{2} \tilde{\sigma}_k.
    \end{array}
\end{equation}
Equation \eqref{eq: beta condition_fading_convex} implies that $-(1-\frac{L\beta}{2}\mu_h) \leq -\frac{1}{2}$. Hence, 
\begin{equation}
    \label{eq: F_k+1_bound_FadingChannel_Final}
    \begin{array}{l}
\displaystyle  \mathbb{E}[F(\boldsymbol{\theta_{k+1}}] \leq  \mathbb{E}[F(\boldsymbol{\theta_k})] -\frac{\beta \mu_h}{2}\mathbb{E}[||\boldsymbol{\nabla} F(\boldsymbol{\theta_k})||^2] +\frac{L\beta^2}{2} \tilde{\sigma_k}.   
    \end{array}
\end{equation}
The convexity of $F(\boldsymbol{\theta})$ yields:
\begin{equation}
\mathbb{E}\left[F(\boldsymbol{\theta_k})\right] \leq F(\boldsymbol{\theta^*}) + \mathbb{E}\left[ \boldsymbol{\nabla} F(\boldsymbol{\theta_k})^T\left( \boldsymbol{\theta}_{k} -\boldsymbol{\theta^*}\right) \right].
\end{equation}
By applying the last inequality into \eqref{eq: F_k+1_bound_FadingChannel_Final}, we have
\begin{equation}
    \begin{array}{l}
        \displaystyle 
         \mathbb{E}[F(\boldsymbol{\theta_{k+1}}] \leq F(\boldsymbol{\theta^*}) + \mathbb{E}\left[ \boldsymbol{\nabla} F(\boldsymbol{\theta_k})^T\left( \boldsymbol{\theta}_{k} -\boldsymbol{\theta^*}\right) \right]
          \vspace{0.2cm}\\ \hspace{2cm}
\displaystyle
         -\frac{\beta \mu_h}{2}\mathbb{E}[||\boldsymbol{\nabla} F(\boldsymbol{\theta_k})||^2] +\frac{L\beta^2}{2} \tilde{\sigma_k},
    \end{array}
\end{equation}
which after some algebraic modifications can also be written as follows:
\begin{equation}
\begin{array}{l}
        \displaystyle 
    \mathbb{E}[F(\boldsymbol{\theta_{k+1}}] - F(\boldsymbol{\theta^*}) \leq \frac{1}{2\beta\mu_h}\mathbb{E} \left[\vphantom{\frac{a}{b}} 2\beta\mu_h \boldsymbol{\nabla} F(\boldsymbol{\theta_k})^T\left( \boldsymbol{\theta}_{k} -\boldsymbol{\theta^*}\right) \right.
         \vspace{0.2cm}\\ \hspace{1.0cm}
\displaystyle
\left. -(\mu_h \beta)^2 ||\boldsymbol{\nabla} F(\boldsymbol{\theta_k})||^2 -||\boldsymbol{\theta_k} -\boldsymbol{\theta^*}||^2 + ||\boldsymbol{\theta_k} -\boldsymbol{\theta^*}||^2 \vphantom{\frac{a}{b}} \right] 
    \vspace{0.2cm}\\ \hspace{6cm}
    \displaystyle
    +\frac{L\beta^2}{2} \tilde{\sigma_k}  
    \vspace{0.2cm}\\ \hspace{0.5cm}
    \displaystyle
    =\frac{1}{2\beta\mu_h} \mathbb{E} \left[ ||\boldsymbol{\theta_k} -\boldsymbol{\theta^*}||^2 - ||\boldsymbol{\theta_k}-\beta\mu_h\boldsymbol{\nabla} F(\boldsymbol{\theta_k})-\boldsymbol{\theta^*}||^2 \right]\vspace{0.2cm}\\ \hspace{6cm}
    \displaystyle
    +\frac{L\beta^2}{2} \tilde{\sigma_k} .
\end{array}
\end{equation}
Let $r_k^2 \triangleq ||\boldsymbol{\theta_k} -\boldsymbol{\theta^*}||^2$ and $\tilde{r}_{k+1}^2 \triangleq ||\boldsymbol{\theta_k}-\beta\mu_h\boldsymbol{\nabla} F(\boldsymbol{\theta_k})-\boldsymbol{\theta^*}||^2$.
By summing over $k$ iterations we get
\begin{equation}
\label{eq:nearly telscopic sum}
    \begin{array}{l}
        \displaystyle
        \sum_{i=1}^k \mathbb{E}[F(\boldsymbol{\theta}_{i}] - F(\boldsymbol{\theta^*}) \leq  \frac{1}{2\beta\mu_h} \sum_{i=1}^k \mathbb{E}\left[ r_{i-1}^2 - \tilde{r}_{i}^2\right] 
        \vspace{0.2cm}\\ \hspace{5cm}
    \displaystyle
        +\frac{L\beta^2}{2}\sum_{i=1}^k\tilde{\sigma_i}.
    \end{array}
\end{equation}
 The difference $\mathbb{E}[r_k^2 - \tilde{r}_k^2]$ is equal to $\beta^2 \tilde{\sigma_k}$. Therefore, the first sum on the RHS of (\ref{eq:nearly telscopic sum}) is nearly telescopic up to the additional term $\beta^2 \tilde{\sigma_k}$ which yields:
 \begin{equation}
     \begin{array}{l}
\displaystyle\sum_{i=1}^k  \mathbb{E}[F(\boldsymbol{\theta}_{i}] - F(\boldsymbol{\theta^*})  \leq  \frac{1}{2\beta\mu_h}\left(r_0^2 -\tilde{r}_k^2 +\sum_{i=2}^{k} \beta^2\tilde{\sigma_i}\right) 
          \vspace{0.2cm}\\ \hspace{6cm}
    \displaystyle
    + \frac{L\beta^2}{2} \sum_{i=1}^k\tilde{\sigma}_i.
     \end{array}
 \end{equation}
 Next, by applying the assumption that  $\mathbb{E}\left[||\boldsymbol{\nabla} f_n(\boldsymbol{\theta})||^2\right]<B(N)$ we get
 \begin{equation}
     \begin{array}{l}
\displaystyle\frac{1}{k}\sum_{i=1}^k \mathbb{E}[F(\boldsymbol{\theta}_{i}] - F(\boldsymbol{\theta^*})     \vspace{0.2cm}\\ \hspace{0.5cm}
    \displaystyle
    \leq \frac{r_0^2}{2\beta\mu_h k}+\frac{\beta}{2}\left(\frac{1}{\mu_h}+L\beta
    \right) \left(\frac{\sigma_h^2 B(N)}{N}+ \frac{d\sigma_w^2}{E_N N^2} \right)
    \vspace{0.2cm}\\ \hspace{0.5cm}
    \displaystyle
    \leq\frac{r_0^2}{2\beta\mu_h k}+\frac{\beta}{\mu_h}\left(\frac{\sigma_h^2B(N)}{N}+ \frac{d\sigma_w^2}{E_N N^2} \right).
     \end{array}
 \end{equation}
Recall that the series $ \{ E[F(\boldsymbol{\theta}_{k}] - F(\boldsymbol{\theta^*}) \}$ is monotonically decreasing for $i=1, ..., k$ by by condition (\ref{eq:condition_decreasing}). Hence,  
\begin{equation}
\begin{array}{l}
\displaystyle
    \mathbb{E}[F(\boldsymbol{\theta}_{k}] - F(\boldsymbol{\theta^*})\leq  \frac{1}{k}  \sum_{i=1}^k  \mathbb{E}[F(\boldsymbol{\theta}_{i}] - F(\boldsymbol{\theta^*}) 
       \vspace{0.2cm} \\ \hspace{1.4cm}
    \displaystyle
   \leq 
   \frac{r_0^2}{2\beta\mu_h k}+\frac{\beta}{\mu_h}\left(\frac{\sigma_h^2B(N)}{N}+ \frac{d\sigma_w^2}{E_N N^2} \right),
     \end{array}
\end{equation}
which completes the proof. 
\hfill $\square$
  
\bibliographystyle{IEEEtran}

\end{document}